\documentclass[11pt,a4paper]{article}
\usepackage{amsmath,amssymb}
\usepackage{graphicx}
\usepackage{booktabs}
\usepackage{multirow}
\usepackage{hyperref}
\usepackage{xcolor}
\usepackage{geometry}
\usepackage{enumitem}
\usepackage{tcolorbox}
\tcbuselibrary{breakable,skins}
\usepackage{caption}
\usepackage{subcaption}
\usepackage{array}
\usepackage{fontspec}
\usepackage[fontset=fandol]{ctex}
\usepackage[numbers,sort&compress]{natbib}
\usepackage{titling}

\newcommand{\ra}[1]{\renewcommand{\arraystretch}{#1}}
\newcolumntype{L}[1]{>{\raggedright\arraybackslash}p{#1}}
\newcolumntype{C}[1]{>{\centering\arraybackslash}p{#1}}

\providecommand{\wordmarkfont}{\rmfamily}
\newcommand{\stepxmark}{{\wordmarkfont\bfseries\fontsize{22pt}{24pt}\selectfont StepX}}
\pretitle{%
  \vspace*{-2.2em}
  \noindent
  \begin{minipage}[b]{0.68\linewidth}%
    \stepxmark%
  \end{minipage}%
  \hfill
  \begin{minipage}[b]{0.26\linewidth}%
    \raggedleft\includegraphics[height=4.2em]{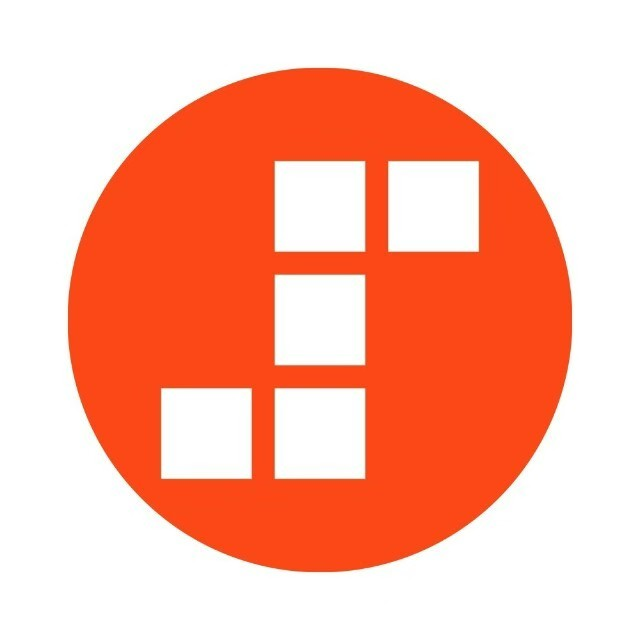}%
  \end{minipage}\par\vspace{0.15em}%
  \noindent\rule{\linewidth}{0.6pt}\par\vspace{3.0em}%
  \begin{center}%
  \Large\bfseries%
}
\posttitle{%
  \par\vspace{0.5em}%
  \end{center}%
  \vspace{0.6em}%
}
\preauthor{\begin{center}\normalsize}
\postauthor{\end{center}}
\predate{\begin{center}\small}
\postdate{\end{center}}

\newtcolorbox{takeawaybox}{
  colback=cyan!4,
  colframe=cyan!40!blue!30,
  fonttitle=\bfseries\color{blue!60!black},
  title=Takeaway,
  boxrule=0.8pt,
  arc=3pt,
  left=8pt,
  right=8pt,
  top=4pt,
  bottom=4pt
}

\newtcolorbox{caseframe}{
  colback=white,
  colframe=black!15,
  arc=3mm,
  boxrule=0.7pt,
  left=8pt, right=8pt, top=8pt, bottom=8pt,
  width=\linewidth,
  breakable=false,
}
\newtcolorbox{promptbox}{
  colback=gray!7,
  colframe=gray!30,
  arc=1.5mm,
  boxrule=0.4pt,
  left=6pt, right=6pt, top=4pt, bottom=4pt,
  fontupper=\small,
}
\newtcolorbox{responsebox}{
  colback=cyan!3,
  colframe=cyan!30!blue!25,
  arc=1.5mm,
  boxrule=0.4pt,
  left=6pt, right=6pt, top=4pt, bottom=4pt,
  fontupper=\small,
}

\title{StepX-Edge: An On-Device UI Vision--Language Model\\via Architecture--Training--Deployment Co-Design}
\author{%
  {\normalsize\bfseries Yin Wang \quad Haotian Hu \quad Jineng Han \quad Wentao Qiu \\[0.3em]
  Zhenhua Ge \quad Liujian Tang \quad Fanyi Wang} \\[0.4em]
  {\small StepX Team}%
}
\date{}

\begin{document}
\maketitle

\begin{abstract}
Deploying vision--language models with full UI-understanding capabilities to on-device settings has long been hindered by a dilemma between accuracy and efficiency: on one side lie the accuracy thresholds for four capabilities---OCR, screen understanding, visual question answering, and element grounding---while on the other stand the stringent constraints that mobile chips impose on compute, memory, and power. Existing work either sacrifices one capability to gain another, or remains at the simulation level without validation on real hardware. We present \textbf{StepX-Edge}, a 0.9B-parameter on-device UI vision--language model that resolves this tension through a three-layer co-design of \textbf{architecture, training, and deployment}. At the architectural level, UI-Aware Layered Visual Encoding (ULVE) and the Progressive Dimensionality Projection (PDP) connector are vertically optimized for the extreme aspect ratios and fine-grained perception demands of screens, and standard full attention is used throughout to ensure native compatibility with mainstream mobile NPU operators. At the training level, the five-stage progressive StepX-Curriculum framework---designed from our observation of the mutually reinforcing effects among UI tasks---enables the four capabilities to grow synergistically rather than interfere with one another under a small parameter budget. At the deployment level, a module-wise, differentiated PTQ$\to$QAT two-stage quantization scheme keeps the post-quantization accuracy loss within 1\%. In experiments, StepX-Edge achieves the strongest overall UI-understanding performance in the $\leq$1B tier with only 0.9B parameters, surpassing all 2B--2.3B baselines on ScreenQA (88.76 F1) and Chinese OCRBench v2 (57.25), and matching 1.3B--2.3B general-purpose VLMs on RefCOCO (92.0\%) and OCRBench v1 (831) with a substantially smaller parameter count. After W4A16+KV8 quantization, the model runs stably and end-to-end usable on a Snapdragon 8 Gen5 device, with a time-to-first-token of about 0.84\,s, a decode speed of 98 tok/s, and a peak memory of 1.4\,GB. We will open-source all training data, the complete training pipeline, and the quantized deployment scheme.
\end{abstract}

\section{Introduction}

Multimodal large language models (MLLMs)~\cite{flamingo,llava,gpt4v,qwenvl,minicpmv46} are rapidly advancing the frontier of artificial intelligence, enabling machines to deeply understand and reason over multimodal information such as text and images. As MLLMs extend into vertical domains, UI understanding has become a core scenario for building agents capable of interacting with graphical user interfaces. However, deploying highly capable screen-understanding models to on-device settings (e.g., smartphones) remains extremely challenging: the strict constraints on compute, memory, and power make the ``accuracy--efficiency'' trade-off the central bottleneck for on-device VLMs, and a focal point of shared interest across both research and industry.

We decompose this efficiency--accuracy problem into three core aspects. \textbf{(1) Model architecture.} UI screens exhibit visual characteristics that differ markedly from natural images: extreme aspect ratios (often as high as 20:9), densely packed text and icons, and fine-grained spatial-structural relationships. These properties demand stronger spatial encoding and fine-grained perception, yet general-purpose small VLMs pretrained on natural images (e.g., MiniCPM-V 4.6~\cite{minicpmv46}, Gemma4-E2B-it~\cite{gemma4}, Qwen3.5-VL-0.8B~\cite{qwen35vl}) carry visual inductive biases mismatched with UI scenarios, so direct transfer incurs noticeable performance degradation. Moreover, many cloud-oriented efficiency designs (e.g., sliding-window attention~\cite{mistral}, sparse attention~\cite{sparse_attention}) are poorly supported by current mobile NPU/GPU operators, introducing additional deployment overhead and engineering risk. \textbf{(2) Training strategy.} UI understanding requires a model to master four capabilities simultaneously: OCR, screen-content understanding, visual question answering (VQA), and UI element grounding. Naively mixing multi-task data for training often causes the capabilities to interfere rather than cooperate---the final performance on each task falls below that of training it in isolation, a phenomenon especially pronounced under a $\leq$1B parameter budget. Most existing work focuses on a single capability, or naively mixes tasks without exploiting the intrinsic relationships among them. \textbf{(3) Deployment pipeline.} Existing on-device VLM work largely centers on accuracy metrics against cloud benchmarks, treating deployment efficiency as an afterthought. Relying solely on post-training quantization (PTQ)~\cite{gptq,awq,smoothquant} often leads to unacceptable accuracy loss on UI tasks---these tasks place stringent demands on precise text recognition and pixel-level grounding, and low-bit quantization errors are amplified token by token during autoregressive decoding. Furthermore, much work claims to ``support on-device deployment'' but remains at the simulation or theoretical-compute level, lacking complete validation on real hardware.

To address these challenges, StepX-Edge introduces three key improvements across model architecture, training strategy, and the deployment pipeline. \textbf{(1) A UI-aware, efficient on-device architecture.} StepX-Edge is a 0.9B-parameter VLM designed specifically for mobile UI understanding, adopting a three-stage architecture: a Qwen3.5 vision encoder~\cite{qwen35vl} ($\sim$0.3B) adapted to the domain via the UI-Aware Layered Visual Encoding strategy (ULVE), a three-layer Progressive Dimensionality Projection connector (PDP, $\sim$6.3M) for efficient vision--language alignment, and Qwen3-0.6B~\cite{qwen3} for semantic understanding and instruction generation. The model uses standard full attention throughout to maximize compatibility with mobile-chip operators and to minimize the engineering cost of migrating from a research prototype to production deployment. \textbf{(2) The five-stage progressive StepX-Curriculum training framework.} We find that OCR, screen understanding, VQA, and grounding exhibit strong mutual reinforcement during joint training---the grounding task drives the model to acquire precise spatial encoding, which in turn improves the positional accuracy of OCR; the textual semantics accumulated by OCR then provide stronger anchor signals for grounding, forming a positive feedback loop. Building on this, we design a five-stage progressive curriculum (modality-alignment pretraining $\to$ multi-task pretraining $\to$ foundational SFT $\to$ advanced SFT $\to$ RL alignment), paired with a data recipe optimized for mobile resolutions and aspect ratios, so that the four capabilities grow synergistically under a 0.9B budget. The entire training requires only about 1{,}536 GPU-hours ($\sim$1 day on 64 GPUs) and roughly 30 billion tokens of multimodal corpus. \textbf{(3) A real-device-validated, module-wise PTQ$\to$QAT deployment scheme.} Targeting the computational characteristics and accuracy sensitivity of each module, we adopt differentiated quantization (W8A16 for the ViT and connector, W4A16+KV8 for the LLM) and a two-stage procedure: PTQ is first applied to the ViT and connector and then frozen, after which QAT is performed on the LLM to compensate for 4-bit quantization error, keeping the accuracy loss relative to the floating-point baseline within 1\%. We complete the full real-device validation---from model conversion and quantization to end-to-end inference---on the Snapdragon 8 Gen5 platform.

Comprehensive experiments show that, with only 0.9B parameters, StepX-Edge achieves the strongest overall UI-understanding performance in the $\leq$1B tier, matching or even surpassing general-purpose VLMs with 1.3B--2.3B parameters on multiple tasks. For instance, on the screen question-answering benchmark ScreenQA it attains an F1 of 88.76---the highest among all baselines---leading the runner-up Qwen3.5-VL-0.8B (75.41) by roughly 13 points; on the Chinese text-recognition benchmark OCRBench v2 (cn) it ranks first among all baselines with 57.25 (the runner-up is Qwen3-VL-2B-Instruct at 53.0). On general grounding and OCR tasks, StepX-Edge attains performance comparable to much larger models with a substantially smaller parameter count: 92.0\% on RefCOCO grounding, surpassing same-tier models and the 1.3B MiniCPM-V 4.6 (86.7) and approaching the 2B Qwen3-VL-2B-Instruct~\cite{qwen3vl} (93.62); and 831 on OCRBench v1, outperforming MiniCPM-V 4.6 (824) and approaching Qwen3-VL-2B-Instruct (859). Notably, these results are achieved under the premise of on-device deployability: after W4A16+KV8 quantization, StepX-Edge runs on a Snapdragon 8 Gen5 device at a decode throughput of about 98 tok/s, a time-to-first-token of about 0.84\,s, and a peak runtime memory of about 1.4\,GB, while its parameter count is only about 39\% of Gemma4-E2B-it (2.3B) and about 69\% of MiniCPM-V 4.6 (1.3B).

In summary, the contributions of this paper are as follows:
\begin{enumerate}[nosep]
    \item We present StepX-Edge, a 0.9B-parameter mobile VLM supporting four capabilities---OCR, screen understanding, VQA, and UI grounding. It attains the highest scores among all baselines on ScreenQA and Chinese OCRBench v2, matches 1.3B--2.3B general-purpose VLMs on benchmarks such as RefCOCO and OCRBench v1 with a smaller parameter count, and can be deployed efficiently on mobile devices.
    \item We discover and validate the mutually reinforcing effects among UI multi-tasks, and building on this, propose the five-stage progressive StepX-Curriculum training framework together with a mobile-oriented data recipe, enabling the four core capabilities to grow synergistically under a small parameter budget.
    \item We provide a complete, real-device-validated on-device quantized deployment scheme that adopts module-wise differentiated granularity and a two-stage PTQ\allowbreak$\to$\allowbreak QAT procedure, balancing accuracy and efficiency. We plan to open-source the training data, the complete training pipeline, and the quantized deployment scheme, offering the community a reproducible recipe for building on-device VLMs.
\end{enumerate}

\section{StepX-Edge: Method Overview}

StepX-Edge is a 0.9B-parameter on-device vision--language model designed specifically for mobile UI understanding. Its overall architecture follows an end-to-end three-stage design comprising three core modules: (1) a \emph{vision encoder}, responsible for extracting multi-scale visual features from UI screens; (2) a \emph{modality adapter}, which achieves efficient alignment between visual features and language semantics via progressive dimensionality projection; and (3) a \emph{language decoder}, which handles tasks such as UI-content understanding, visual question answering, and instruction generation. To ensure hardware compatibility for on-device deployment, the model uses standard full attention throughout, avoiding operator designs such as sliding-window or sparse attention that are poorly supported by mobile chips.

On the training side, we propose the progressive StepX-Curriculum framework, which fully exploits the mutual reinforcement among the four tasks---OCR, screen understanding, VQA, and UI grounding---to let the four capabilities grow synergistically under a 0.9B parameter budget. On the deployment side, we design a module-wise differentiated two-stage PTQ$\to$QAT quantization pipeline that enables efficient on-device inference with a 4-bit language model while minimizing accuracy loss.

This chapter introduces, in turn, the overall architecture of StepX-Edge (\S\ref{sec:arch}), the data engineering and curriculum recipe (\S\ref{sec:data}), the StepX-Curriculum training framework (\S\ref{sec:training}), and the on-device quantized deployment scheme (\S\ref{sec:deploy}).

\subsection{Model Architecture}\label{sec:arch}

\begin{figure}[!t]
    \centering
    \includegraphics[width=\linewidth]{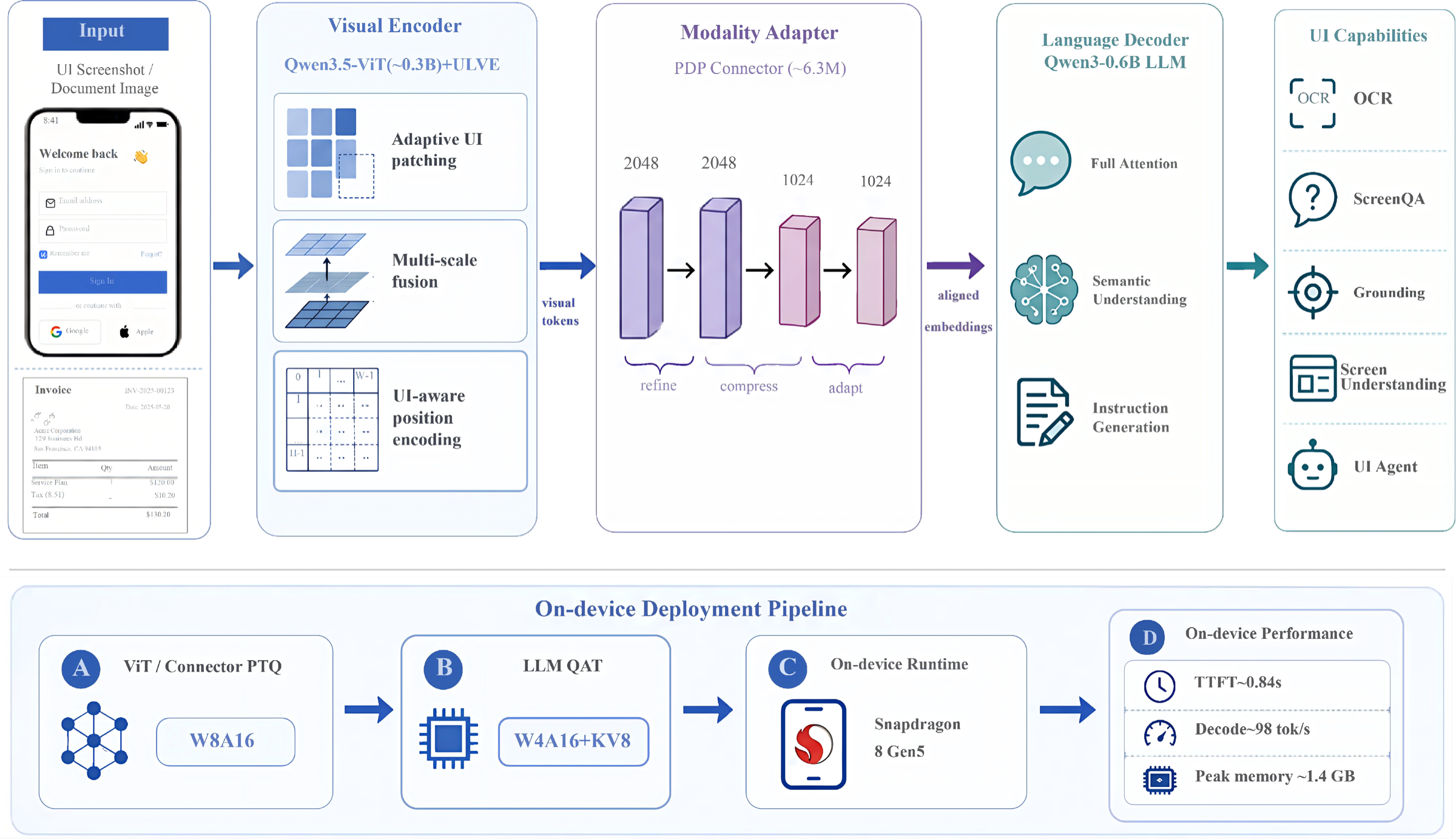}
    \caption{Overall architecture of StepX-Edge and its on-device deployment pipeline. \textbf{Top:} the three-stage architecture---vision encoder (Qwen3.5-ViT + ULVE, $\sim$0.3B) $\to$ PDP connector ($\sim$6.3M) $\to$ language decoder (Qwen3-0.6B, full attention)---which uniformly supports OCR, screen understanding, VQA, and grounding. \textbf{Bottom:} the on-device deployment pipeline---ViT/Connector PTQ (W8A16) $\to$ LLM QAT (W4A16+KV8) $\to$ Snapdragon 8 Gen5 device (TTFT $\sim$0.84\,s, $\sim$98 tok/s, peak memory $\sim$1.4\,GB).}
    \label{fig:arch}
\end{figure}

As shown in Figure~\ref{fig:arch}, StepX-Edge adopts a vision-encoder--modality-adapter--language-decoder three-stage end-to-end multimodal architecture, with a total of about 0.9B parameters. The overall design follows a \textbf{``deployment-friendly first''} engineering paradigm: while preserving UI-understanding capability, it prioritizes structural choices with high operator maturity, good hardware compatibility, and strong cross-platform consistency, so as to minimize the engineering cost of migrating from a research prototype to production-grade deployment.

The vision encoder is built on the pretrained Qwen3.5-ViT, which we systematically adapt via the \textbf{UI-Aware Layered Visual Encoding strategy (ULVE)}; the modality adapter adopts a \textbf{Progressive Dimensionality Projection (PDP)} design that achieves efficient cross-modal alignment while remaining lightweight; and the language decoder is built on Qwen3-0.6B, handling semantic understanding and instruction generation. The model uses standard full attention throughout to ensure maximal compatibility with mainstream mobile NPU/GPU operator libraries. The parameter distribution across components is shown in Table~\ref{tab:arch}.

\begin{table}[!htbp]
\centering
\small
\caption{Parameter distribution across StepX-Edge components.}\label{tab:arch}
\ra{1.25}
\begin{tabular}{@{}l L{8.2cm} r@{}}
\toprule
\textbf{Component} & \textbf{Core technique} & \textbf{Parameters} \\
\midrule
Vision encoder    & Qwen3.5-ViT + ULVE adaptation                                       & $\sim$0.3B \\
Modality adapter  & Three-layer progressive dimensionality projection MLP (PDP)         & $\sim$6.3M \\
Language decoder  & Qwen3-0.6B + full attention                                         & 0.6B \\
\midrule
\multicolumn{2}{@{}l}{\textbf{Total}} & $\sim$\textbf{0.9B} \\
\bottomrule
\end{tabular}
\end{table}

\subsubsection{UI-Aware Layered Visual Encoding (ULVE)}

The vision encoder is the perceptual foundation of UI-understanding capability. General-purpose ViTs are pretrained primarily on natural images, and their feature-extraction patterns exhibit a significant domain gap from the visual characteristics of UI scenarios---UI screens have unique properties such as extreme aspect ratios, densely packed text and icons, and precise spatial-layout relationships. To bridge this domain gap, we propose the \textbf{UI-Aware Layered Visual Encoding (ULVE)} strategy, which systematically adapts Qwen3.5-ViT at three levels: tiling strategy, feature fusion, and training mechanism.

\textbf{(1) Adaptive tiling and positional-encoding optimization for extreme aspect ratios.} The tiling strategy of a general ViT is typically designed for near-square natural images, whereas UI screens often have aspect ratios as high as 20:9 or even more extreme. Directly applying generic tiling over-compresses the long dimension and over-splits the short dimension, distorting and losing spatial information. Inspired by the ``any-aspect-ratio-aware'' idea in LLaVA-UHD~\cite{llava_uhd}, we perform UI-specific optimization of the tiling strategy: given the actual aspect ratio $r = H:W$ of the input screen, and subject to a budget on the total number of tiles, we select the row-column decomposition $h \times w$ that makes each tile's aspect ratio closest to 1:1 (i.e., minimizing $|(H/h)/(W/w) - 1|$), thereby ensuring a uniform distribution of visual features within each tile. Meanwhile, we correspondingly recalibrate the 2D positional encoding so that the model can more precisely perceive the absolute positions and relative relationships of UI elements within elongated screens. This design specifically mitigates the spatial-information distortion in long-interface understanding and small-element grounding on portrait screens.

\textbf{(2) A UI-optimized multi-scale feature weighted-fusion mechanism.} UI understanding relies simultaneously on fine-grained low-level features (e.g., text strokes, icon details) and global high-level features (e.g., interface layout, structural relationships). The feature fusion of a general Merger layer is uniformly weighted, yet the UI scenario has unbalanced demands across feature levels---text recognition relies more on low-level features, while layout understanding relies more on high-level features. We introduce a channel-attention module after the PixelShuffle~\cite{pixelshuffle} operation in the Merger layer, applying learnable adaptive weighting to feature maps from different levels. Denoting the features from each level as $F_1,\ldots,F_L$, the channel attention produces weights $\alpha = \mathrm{softmax}(g(F))$, and the fused feature is
\begin{equation}
\tilde{F} = \sum_{l=1}^{L} \alpha_l \cdot F_l,
\end{equation}
where $g(\cdot)$ is a lightweight channel-scoring network. This enables the model to dynamically ``attend'' to the more important feature levels according to the current task, rather than weighting all levels uniformly.

\textbf{(3) An early domain-adaptation mechanism with a fully trainable Merger layer.} The traditional transfer-learning paradigm usually proceeds by ``freezing lower layers first, then progressively unfreezing.'' However, we find that for UI understanding, the Merger layer---as the high-level aggregator of visual features---has a speed and quality of domain adaptation that directly determine the ceiling of subsequent modality alignment. We therefore adopt a strategy in which \textbf{the Merger layer is trainable throughout}: from the modality-alignment pretraining of Stage 1, the Merger layer is trained together with the Connector, learning early how to reshape pretrained visual features into a form better suited to UI scenarios and cross-modal alignment; meanwhile, the main Transformer layers of the ViT remain frozen until Stage 3 (foundational-capability SFT), where they are progressively unfrozen (fine-tuned at 1/10 of the LLM learning rate). This ``adapt the high level first, fine-tune the low level later'' layered strategy leverages the high plasticity of the Merger layer to complete domain adaptation quickly, while freezing the main layers preserves the general visual knowledge from pretraining, avoiding damage to low-level features by early-training noise.

\begin{takeawaybox}
When performing domain adaptation on a general ViT, make the high-level aggregator (Merger) trainable early and throughout, while freezing the low-level Transformer---the high level is highly plastic and adapts quickly, whereas freezing the low level safeguards the general visual knowledge from pretraining.
\end{takeawaybox}

\subsubsection{Progressive Dimensionality Projection Modality Adapter (PDP)}

In the architectural design of on-device VLMs, the modality connector is the key bridge linking vision and language, and its design directly affects the model's final performance and inference efficiency. Existing work typically adopts a single-layer MLP or cross-attention as the connector---the former is structurally simple but incurs large information loss, whereas the latter has strong fusion capability but high computational overhead and poor hardware friendliness. To balance performance and efficiency in the on-device setting, we propose the \textbf{Progressive Dimensionality Projection (PDP)} modality-adapter design.

\textbf{Design motivation.} There is a significant gap between the feature dimension output by the vision encoder (2048-dim) and the embedding dimension of the language model (1024-dim). Directly using a single-layer linear projection for one-shot dimensionality reduction causes a severe information-bottleneck effect---a large amount of task-relevant fine-grained visual information is lost during dimensional compression, which is especially detrimental to tasks that rely on fine-grained features, such as OCR text recognition and UI element grounding. Complex structures such as cross-attention, though theoretically offering stronger fusion capability, significantly increase inference latency and deployment difficulty, running counter to the needs of the on-device setting.

\textbf{Concrete design.} We adopt a three-layer MLP with the dimension configuration $2048 \to 2048 \to 1024 \to 1024$:
\begin{itemize}[nosep]
    \item \textbf{The first layer} is a source-space refinement layer: it applies a nonlinear transformation within the 2048-dim visual space, performing semantic refinement and reorganization of visual features to select task-relevant information and suppress redundancy ahead of the subsequent reduction;
    \item \textbf{The second layer} is a dimensional-transition layer: it performs the key reduction from 2048-dim to 1024-dim, achieving dimensional matching of the feature space with controllable information loss;
    \item \textbf{The third layer} is a target-domain adaptation layer: it further optimizes the feature representation within the 1024-dim language-embedding space, so that the visual representation better matches the semantic distribution of the language model and improves the quality of cross-modal fusion.
\end{itemize}

The core idea of this design is not ``multiple reductions'' but ``refine before reduction, adapt after reduction'': the only substantive dimensional compression (the second layer) is sandwiched between a source-space refinement layer and a target-space adaptation layer. Before reduction, task-relevant information is reorganized and gathered in the original dimension, so that compression happens on an ``already-prepared'' feature space, reducing the loss of fine-grained visual information during compression; after reduction, domain adaptation is performed in the target space so that the compressed representation conforms to the language semantic distribution.

\textbf{Efficiency and hardware friendliness.} Because it uses a pure MLP structure throughout, PDP fully preserves hardware friendliness---its inference latency is comparable to that of a single-layer MLP, far lower than cross-attention-style structures, and it is natively supported by all mainstream mobile inference frameworks without extra operator adaptation. In addition, the three-layer MLP has a total of only about 6.3M parameters, accounting for less than 1\% of the 0.9B total budget---a highly cost-effective, lightweight module.

\begin{takeawaybox}
An on-device connector need not rely on heavier structures (such as cross-attention); within a pure MLP, the division of labor of ``refine before reduction, adapt after reduction'' can mitigate the information bottleneck of one-shot reduction at extremely low parameter and operator cost.
\end{takeawaybox}

\subsubsection{Language Decoder}

The language decoder is built on Qwen3-0.6B~\cite{qwen3}, with a hidden dimension of 1024, and undertakes the core functions of multimodal semantic understanding, reasoning, and instruction generation. It receives the aligned visual representation and text instructions and performs diverse tasks such as UI-interface summarization, visual question answering, element-grounding description, and multi-turn dialogue. Two key technical trade-offs underlie the selection and design of the language decoder:

\textbf{(1) A Pareto-optimal choice of parameter count.} In the on-device deployment setting, there is a clear trade-off between the language model's parameter count and its inference performance, memory footprint, and latency. We conducted a systematic selection comparison across the 0.5B--1B parameter range and chose 0.6B as the balance point under current hardware conditions---it retains sufficient semantic-understanding and reasoning capability to handle complex UI-interaction demands, while keeping the memory footprint within the comfortable range of mainstream mobile devices after 4-bit quantization. Moreover, Qwen3-0.6B's performance in Chinese understanding, instruction following, and structured output aligns closely with the task requirements of UI agents.

\textbf{(2) Deployment-engineering considerations of full attention.} The model uses standard full attention throughout~\cite{transformer}. In recent years, techniques such as sliding-window attention and sparse attention have been shown theoretically to reduce the computational complexity of long-sequence inference; however, in actual on-device deployment, these special attention patterns commonly suffer from insufficient operator maturity, inadequate hardware optimization, and poor cross-platform consistency, often requiring substantial engineering adaptation to reach the expected performance. By contrast, standard full attention is the most thoroughly optimized and most widely supported attention pattern in current mobile-chip operator libraries, with the lowest deployment cost and the least landing risk. For UI-understanding scenarios, the number of input visual tokens is within a controllable range, and the computational overhead of full attention is within an acceptable threshold; full attention is therefore the optimal solution between engineering efficiency and runtime efficiency.

\vspace{0.5em}
\noindent\textbf{Architecture design summary.} The architectural philosophy of StepX-Edge can be summarized as ``deeply optimize the known, cautiously explore the unknown'': for core capabilities such as visual encoding, modality alignment, and language decoding, we perform deep adaptation and progressive optimization on top of mature pretrained models (ULVE improves the UI adaptability of visual perception; PDP improves the efficiency of cross-modal alignment); for deployment-related engineering decisions, we prioritize solutions that have been validated at large scale, rather than pursuing theoretical sophistication.

\begin{takeawaybox}
In architecture design for on-device settings, performing vertical-domain adaptation on mature pretrained modules and confining structural innovation to the range of verifiable benefit often balances performance and deployability better than introducing cutting-edge designs whose operators are not yet mature.
\end{takeawaybox}

\subsection{Data Engineering and Curriculum Recipe}\label{sec:data}

Data is the decisive factor for on-device UI-understanding capability. The training data of general multimodal models is dominated by natural images, in which UI scenarios are underrepresented and unevenly distributed, so direct transfer often works poorly. StepX-Edge builds a large-scale training corpus---mainly from public datasets and covering the four core capabilities---and organizes it through a three-stage curriculum mixing scheme; the first three large-scale training stages use about 10.5M samples in total, covering hundreds of heterogeneous data sources. The data scale, composition, and objective of each stage are shown in Table~\ref{tab:data}.

\begin{table}[!htbp]
\centering
\small
\caption{The three-stage training data recipe of StepX-Edge.}\label{tab:data}
\ra{1.25}
\begin{tabular}{@{}l c c L{6.8cm}@{}}
\toprule
\textbf{Stage} & \textbf{Samples} & \textbf{Sources} & \textbf{Stage objective} \\
\midrule
Stage-PT (projector pretraining)   & $\sim$3.07M & 305 & Modality alignment; establish general visual semantics \\
Mid-training (multi-task)          & $\sim$4.95M & 64  & Inject four capabilities; shift focus toward UI \\
SFT (supervised fine-tuning)       & $\sim$2.56M & 63  & Refine four capabilities; improve accuracy \\
\bottomrule
\end{tabular}
\end{table}

As shown in Table~\ref{tab:data}, the first three training stages constitute the large-scale training body of StepX-Edge; their mapping to the five-stage curriculum in \S\ref{sec:training} is as follows: Stage-PT corresponds to Stages 1--2 of the curriculum (modality alignment and multi-task pretraining), while the Mid-training and SFT data serve, respectively, the multi-task pretraining and foundational-capability fine-tuning of Stages 2--3. The later curriculum stages---Stage 4 (advanced-capability SFT) and Stage 5 (RL alignment)---follow a ``small, carefully selected'' route, and their data composition and reward design are detailed in \S\ref{sec:training}; this chapter focuses on the already-finalized large-scale training data of the first three stages.

\subsubsection{Data Sources and Composition}

The training data is dominated by public datasets, organized around the four core capabilities and a general base:
\begin{itemize}[nosep]
    \item \textbf{OCR and documents}: scene text (LSVT~\cite{lsvt}, MTWI~\cite{mtwi}, COCO-Text~\cite{coco_text}, SynthText~\cite{synthtext}, TextOCR~\cite{textocr}), documents and receipts (SROIE~\cite{sroie}, XFUND~\cite{xfund}, FUNSD~\cite{funsd}, PubTabNet~\cite{pubtabnet}, wildreceipt~\cite{wildreceipt}), handwriting (CASIA-HWDB~\cite{casia_hwdb}), and large-scale printed and synthetic text (MJSynth~\cite{mjsynth}, IDL-WDS~\cite{idl_wds}, HierText~\cite{hiertext}). This data is used after text-box coordinate validation and recognition-consistency filtering, to strengthen full-spectrum recognition and structured parsing from street-scene text to dense documents.
    \item \textbf{UI grounding and understanding}: mobile and web UI data (SeeClick~\cite{seeclick}, ShowUI-Web~\cite{showui}, WaveUI-25k~\cite{waveui}, MobileViews~\cite{mobileviews}, Screen2Words~\cite{screen2words}) and real-device interaction trajectories (Android Control~\cite{android_control}, AITW~\cite{aitw}, AMEX~\cite{amex}, GUIAct~\cite{guiact}, AndroidWorld~\cite{androidworld}, plus in-house real-app screenshots). All samples are uniformly aligned to mobile resolutions and aspect ratios, and bounding boxes are checked for normalized-range validity, providing supervision for element referring, grounding, and interface semantics.
    \item \textbf{Visual question answering and reasoning}: screen and document QA (ScreenQA~\cite{screenqa}, DocVQA~\cite{docvqa}, InfographicVQA~\cite{infographicvqa}, ChineseDocVQA~\cite{chinesedocvqa}), chart and text VQA (ChartQA~\cite{chartqa}, TextVQA~\cite{textvqa}, ST-VQA~\cite{stvqa}, OCR-VQA~\cite{ocrvqa}), and general reasoning (ScienceQA~\cite{scienceqa}, MathVista~\cite{mathvista}). This data is used after filtering QA pairs for linguistic validity and length distribution, to enhance multi-step reasoning and semantic understanding.
    \item \textbf{General image-text (to retain base capability)}: large-scale Chinese and English image-text pairs (LAION~\cite{laion}, CC12M~\cite{cc12m}, Wukong-100M~\cite{wukong}, COCO-CN~\cite{coco_cn}, Flickr30k-CN~\cite{flickr30k_cn}) and high-quality instruction data (ALLaVA-4V~\cite{allava}, ShareGPT4V~\cite{sharegpt4v}, ShareGPT-4o~\cite{sharegpt4o}, SVIT~\cite{svit}, the LLaVA series~\cite{llava}), mixed in at a lower weight after CLIP~\cite{clip} image-text relevance filtering and perceptual-hash deduplication, to prevent the UI-specialized training from degrading general visual capability.
\end{itemize}

The complete list of data sources is released together with the open-sourced training recipe.

\subsubsection{Curriculum Mixing Strategy}

The data supply of StepX-Curriculum follows three orthogonal curriculum principles that advance in tandem with the training stages:

\textbf{(1) A general$\to$vertical distribution shift.} The sampling weight of general image-text data decreases stage by stage: it participates at full weight (1.0) in Stage-PT to lay a broad visual-semantic foundation; upon entering Mid-training it drops to 0.2--0.3 (e.g., ShareGPT4V, ALLaVA-4V, and LLaVA-ReCap are all 0.3), yielding the training focus to UI and OCR tasks; and it is further reduced in the SFT stage. This shift expands UI-specialized capability while avoiding the ``washing out'' of general visual knowledge.

\textbf{(2) A coarse-to-fine quality grading.} The same data source is split by quality into a mid-quality tier (mid) and a high-quality tier (sft): Mid-training uses the mid-quality tier to maximize coverage and scale, while the SFT stage uses only the rigorously screened high-quality tier to improve accuracy and stability---the SFT data scale (2.56M) is markedly smaller than that of Mid-training (4.95M), but each sample carries higher training value.

\textbf{(3) An engine-task-first weighting.} Based on the discovery of the mutual reinforcement among UI multi-tasks, we let the ``engine'' grounding task ramp up first: in Mid-training we assign relatively high weights to grounding data (e.g., cn\_mobile\_grounding 0.25, seeclick\_hard 0.2), prioritizing the model's acquisition of precise spatial encoding; once the spatial foundation is established, the SFT stage raises OCR, VQA, and grounding to full main weights (mostly 1.0) to strengthen textual semantics and reasoning, which in turn reinforces grounding. The overall sampling-weight range is 0.1--1.0, determined through multiple rounds of iteration.

\subsubsection{Data Cleaning and Quality Control}

Raw heterogeneous data varies widely in quality---real screenshots contain blur, occlusion, and duplication; synthetic data may exhibit layout distortion; and automatic annotation inevitably introduces recognition and alignment errors. To ensure the reliability of training data at the tens-of-millions scale, we design a cascaded data quality-control pipeline through which data must pass stage by stage to enter the training pool; failing any stage results in removal or downgrading:
\begin{itemize}[nosep]
    \item \textbf{Image-level inspection}: detecting and filtering out blurry images based on Laplacian variance, removing over-/under-exposed samples based on brightness histograms, and filtering images with excessively low resolution (e.g., short side $<$ 224px) or abnormal aspect ratios (e.g., $>$ 30:9). This stage mainly cleans low-quality images among real screenshots and web image-text data.
    \item \textbf{Annotation-level validation}: applying structured validation by task type---OCR data is checked for out-of-bounds text-box coordinates and empty text; grounding data is checked for consistency between the normalized bounding-box range and image dimensions, removing boxes that are out of bounds or degenerate (zero width/height); and VQA/caption data is checked for the linguistic validity and length distribution of QA pairs, filtering truncated, garbled, and language-mismatched samples.
    \item \textbf{Cross-source deduplication and leakage prevention}: employing two-level deduplication---on the image side, perceptual hashing (pHash)~\cite{phash} computes fingerprints, and those with a Hamming distance below a threshold are deemed near-duplicates, keeping the highest-quality copy; on the text side, n-gram MinHash~\cite{minhash} estimates Jaccard similarity to deduplicate highly similar annotations; the overall deduplication rate is about 12\%--18\%. To eliminate data leakage, the test-set images of all evaluation benchmarks (OCRBench, RefCOCO, ScreenQA, etc.) are cross-checked against the training pool via the same pHash pipeline before training and removed.
    \item \textbf{Model-assisted re-verification}: for the automatically annotated OCR/document data that constitutes the largest share, we build a reasoning-based re-verification stage powered by a large model (think\_cleaned)---a stronger teacher VLM performs field-by-field consistency checks on the recognition results, and samples with confidence below a threshold enter manual spot-checking or are discarded outright, reducing the field-level error rate while avoiding ``hallucinated'' annotations induced by over-augmentation.
    \item \textbf{Quality grading and admission}: data that passes the above stages is graded by a composite quality score (a weighted combination of annotation confidence, image clarity, task difficulty, and source credibility) into the mid and sft tiers, serving Mid-training and fine-tuning respectively, forming a ``coarse-to-fine'' curriculum supply.
\end{itemize}

Through the above pipeline, the overall elimination rate of raw data is about 30\%--40\%, and the samples entering the training pool reach a controllable level in image quality, annotation correctness, and distributional balance.

\begin{takeawaybox}
The value of data engineering for small on-device models lies not in the number of datasets but in the rigor of quality control and the curriculum-based mixing---through cascaded filtering at the image/annotation/deduplication/model-review levels, combined with the mixing principles of ``general$\to$vertical shift, coarse$\to$fine grading, and engine-task-first,'' a corpus of limited scale can maximize the synergistic gains of multi-tasks under a small parameter budget.
\end{takeawaybox}

\subsection{StepX-Curriculum: A Progressive Training Pipeline}\label{sec:training}

StepX-Edge adopts the progressive training pipeline StepX-Curriculum, achieving the synergistic improvement of four capabilities---OCR, screen understanding, VQA, and UI grounding---under a 0.9B parameter budget. The core of the training design rests on the discovery of the mutual reinforcement among UI multi-tasks: the grounding task improves the model's spatial-encoding capability, which in turn promotes OCR accuracy; the textual-semantic knowledge accumulated by the OCR task, in turn, provides stronger semantic anchors for grounding. Through a well-designed curriculum, the four capabilities form a positive feedback loop rather than interfering with one another.

\begin{figure}[!t]
    \centering
    \includegraphics[width=\linewidth]{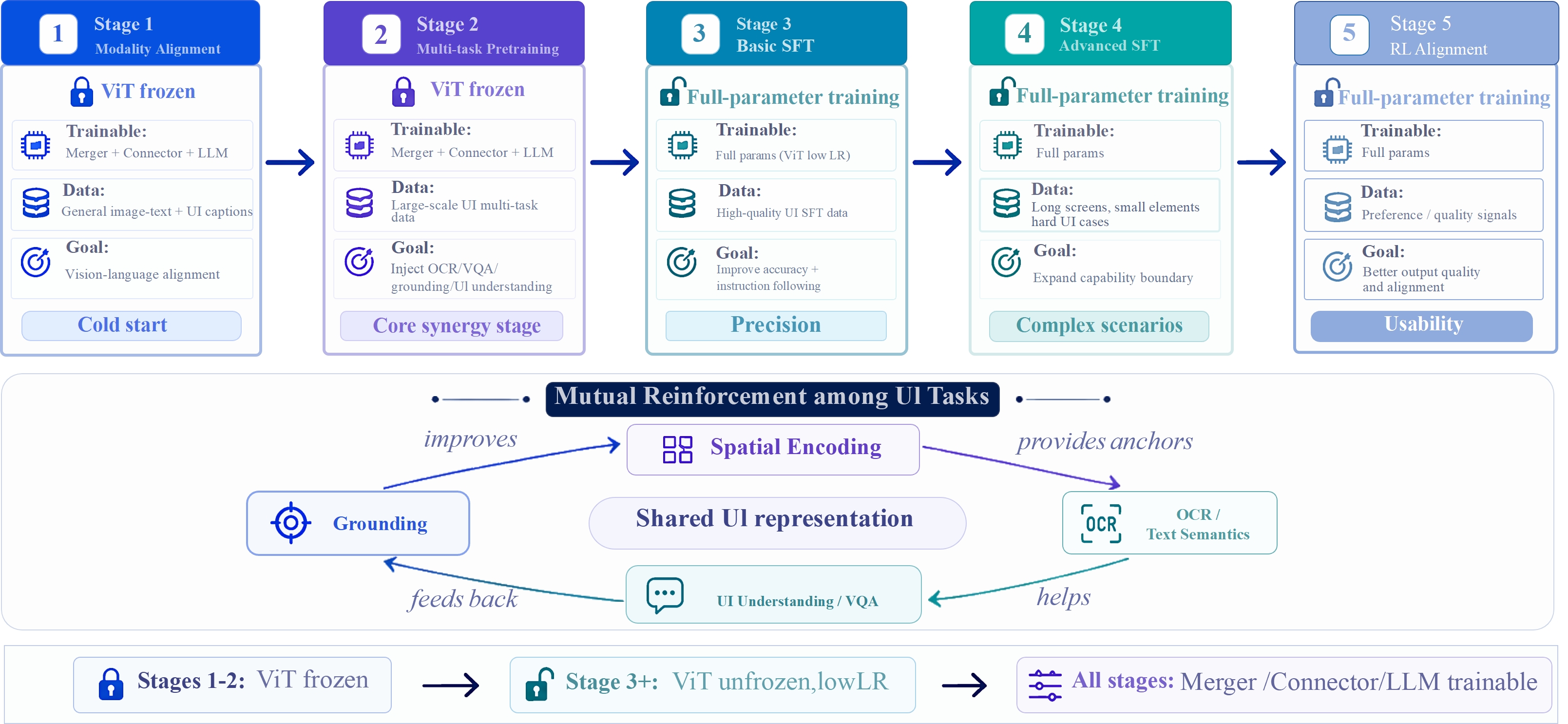}
    \caption{The five-stage progressive training pipeline of StepX-Curriculum. Top: the trainable modules, data, and objectives of Stages 1--5 (cold start $\to$ core mutual reinforcement $\to$ refinement $\to$ complex scenarios $\to$ usability alignment). Middle: the three task groups---OCR, grounding, and UI/VQA---form a positive feedback loop through a shared UI representation. Bottom: Stages 1--2 freeze the ViT, and from Stage 3 onward it is unfrozen at 1/10 of the LLM learning rate.}
    \label{fig:train}
\end{figure}

The pipeline comprises five stages: (i) modality-alignment pretraining, (ii) multi-task pretraining, (iii) foundational-capability supervised fine-tuning, (iv) advanced-capability supervised fine-tuning, and (v) reinforcement-learning alignment. The first three stages constitute the large-scale training body (corresponding to the three data recipes of \S\ref{sec:data}), while the latter two further extend complex-scenario capability and output alignment on top of it. The overall flow of the five stages and the mutual reinforcement among tasks are illustrated in Figure~\ref{fig:train}, and the trainable parameters, data configuration, and training objectives of each stage are shown in Table~\ref{tab:curriculum}.

\begin{table}[!htbp]
\centering
\scriptsize
\caption{Training objectives and data configuration of the five StepX-Curriculum stages.}\label{tab:curriculum}
\ra{1.3}
\setlength{\tabcolsep}{3pt}
\begin{tabular}{@{}L{4.1cm} L{3.3cm} L{4.5cm} L{3.6cm}@{}}
\toprule
\textbf{Training stage} & \textbf{Trainable params} & \textbf{Data configuration} & \textbf{Core objective} \\
\midrule
Stage 1: Modality alignment       & Merger + Connector + LLM  & General image--text + basic UI caption & Basic vision--language alignment \\
Stage 2: Multi-task pretraining   & Merger + Connector + LLM  & Large-scale UI multi-task data         & Jointly inject four basic capabilities \\
Stage 3: Foundational SFT         & Full params (ViT low LR)  & High-quality UI basic tasks            & Consolidate core-capability accuracy \\
Stage 4: Advanced SFT             & Full params               & Complex-scenario UI data               & Extend capability boundaries \\
Stage 5: RL alignment             & Full params               & Preference data                        & Optimize output quality \\
\bottomrule
\end{tabular}
\end{table}

The key training configuration of each stage is shown in Table~\ref{tab:trainconfig}. The first two pretraining stages bear the vast majority of the token budget (about 80\% combined); the SFT stages, though small in sample size, use lower learning rates and aim at refinement. The full pipeline totals about 30 billion tokens and about 1{,}536 GPU-hours (64$\times$A100, about 1 day).

\begin{table}[!htbp]
\centering
\scriptsize
\caption{Hyperparameters and token budget across StepX-Curriculum stages. ``n/a'' marks columns that do not apply to the RL stage (RL uses preference pairs, not a token budget, and does not mix general data).}\label{tab:trainconfig}
\ra{1.3}
\setlength{\tabcolsep}{3pt}
\begin{tabular}{@{}L{3.4cm} L{3.2cm} C{2.0cm} C{1.4cm} L{3.0cm} L{2.4cm}@{}}
\toprule
\textbf{Stage} & \textbf{Trainable modules} & \textbf{Peak LR} & \textbf{Token share} & \textbf{General-data retention} & \textbf{Notes} \\
\midrule
Stage 1: Modality alignment    & Merger + Connector + LLM & 1e-4              & $\sim$35\% & General$:$UI $\approx$ 7$:$3         & ViT frozen \\
Stage 2: Multi-task PT         & Merger + Connector + LLM & 1e-4              & $\sim$45\% & General reduced to $\sim$20\%        & ViT frozen \\
Stage 3: Foundational SFT      & Full params              & 2e-5 (ViT 1/10)   & $\sim$15\% & Retain $\sim$10--15\% general instr. & ViT unfrozen \\
Stage 4: Advanced SFT          & Full params              & 1e-5              & $\sim$5\%  & Small supplement                     & Hard-case focused \\
Stage 5: RL alignment          & Full params              & 5e-6              & n/a        & n/a                                  & GRPO + three-part reward \\
\bottomrule
\end{tabular}
\end{table}

Learning rates are the peak values of each stage, using a warmup + cosine-decay schedule~\cite{cosine_lr}; from Stage 3 the ViT is fine-tuned at 1/10 of the LLM learning rate to balance domain adaptation and the retention of general visual knowledge; the general-data retention ratio decreases stage by stage, which is the concrete realization of the ``general$\to$vertical shift'' curriculum principle of \S\ref{sec:data}.

\subsubsection{Stage 1: Modality-Alignment Pretraining}

\textbf{Objective.} Establish the basic correspondence between visual features and language semantics, complete the cold start of cross-modal alignment, and lay the foundation for subsequent UI-specialized training.

\textbf{Trainable parameters.} The Merger layer + Connector + the full set of LLM parameters participate in training, while the main body of the vision encoder remains frozen.

\textbf{Data.} A mix of general image-text pairs and basic UI image-text pairs, with general caption data (LAION, CC12M, Wukong-100M, COCO-CN, etc.) providing broad semantic coverage and UI caption data (caption versions of Android World/Control, GUI annotations, etc.) introducing preliminary UI-domain features and interface concepts.

\textbf{Design and motivation.} Modality alignment is the foundation of multimodal training and also the stage most prone to training instability. We adopt a ``freeze vision, tune language'' strategy: freezing the main body of the vision encoder preserves the general visual knowledge the pretrained model acquired on large-scale natural images, avoiding damage to low-level visual representations by early-training noise; the LLM participates in training because the language side adapts to multimodal inputs faster, and its early participation accelerates cross-modal semantic alignment; the Merger layer and Connector, as the bridge between vision and language, are the training focus of this stage. The core of this stage is ``speed''---to quickly build the basic vision--language bridge at low cost, transforming the model from a ``pure language model'' into a ``multimodal model that can understand images.''

\subsubsection{Stage 2: Multi-Task Pretraining}

\textbf{Objective.} Exploit the mutual reinforcement among tasks to jointly inject basic versions of the four core capabilities, achieving an overall improvement in UI perception. This is the most critical stage of StepX-Curriculum and the core period in which the mutual-reinforcement effect takes hold.

\textbf{Trainable parameters.} The Merger layer + Connector + the full set of LLM parameters continue training, while the main body of the vision encoder still remains frozen.

\textbf{Data.} Large-scale UI multi-task data covering the four major task categories---OCR, screen understanding, VQA, and UI grounding---mixed according to the curriculum weights of \S\ref{sec:data}. All data is aligned to mobile screen resolutions and aspect ratios, ensuring that the visual distribution of the training data matches the actual deployment scenario.

\textbf{Design and motivation.} The design of this stage stems from our discovery of the mutual-reinforcement effect among tasks. Under the constraint that the main body of the vision encoder is frozen, the model must handle four task demands simultaneously by adjusting the Merger, Connector, and LLM. This ``bottleneck constraint'' appears to be a limitation but is in fact the key to eliciting mutual reinforcement---because the ``entrance'' of the visual representation is fixed, the model cannot learn a separate set of visual features for each task, and can only search within the shared feature space for a general representation that supports multiple tasks simultaneously. Learning under this constraint precisely gives rise to mutual reinforcement: to do grounding well, the model must learn precise spatial encoding, and improved spatial encoding naturally benefits OCR (text-line segmentation and character localization both require spatial perception); to do OCR well, the model must accumulate rich textual-semantic knowledge, and textual semantics in turn serve as strong anchor signals for the grounding task, improving element-grounding accuracy. The four tasks reinforce one another in the shared feature space, forming a positive feedback loop.

\subsubsection{Stage 3: Foundational-Capability Supervised Fine-Tuning}

\textbf{Objective.} Consolidate the accuracy and stability of the four core capabilities on high-quality data, and activate the model's instruction-following capability---aligning the general perceptual capability accumulated during pretraining to diverse human-instruction phrasings, so that the model can reliably respond to real users' UI-task requests posed in different wordings and formats.

\textbf{Trainable parameters.} Entering the full-parameter training stage, the vision encoder is fully unfrozen from this stage on (including the Merger layer), but the ViT learning rate is set to 1/10 of the LLM's.

\textbf{Data.} Carefully selected high-quality UI basic-task data (the sft quality tier), covering the basic scenarios of the four core capabilities. On this basis, we deliberately increase the diversity of instruction phrasing: the same task is presented with multiple instruction templates, multiple phrasings, and multiple output formats (for example, the grounding task includes both natural-language instructions like ``click the login button'' and structured coordinate queries; the OCR task requires both full-screen transcription and region-/field-wise extraction), so that the model learns to understand instructions from semantic intent rather than fixed syntax. Meanwhile, to prevent UI-specialized training from weakening the model's general instruction-following and language capabilities, the training mix retains about 10--15\% of general multimodal instruction data (ALLaVA-4V, ShareGPT4V, the LLaVA series, etc.).

\textbf{Design and motivation.} After the first two stages, the model already possesses the basic capability for UI understanding, but its accuracy, stability, and instruction generalization still need improvement. This stage introduces full-parameter fine-tuning, letting all parameters adjust for UI tasks; the ViT uses a low learning rate to balance domain adaptation and knowledge retention. Instruction diversity is the key design of this stage: in real scenarios, users do not ask questions with the fixed templates used during training, and if the SFT data's instructions are too uniform, the model easily overfits to specific syntax and fails when the phrasing changes. By introducing diverse instruction phrasings and output formats on high-quality data, the model truly ``connects'' the four capabilities to a flexible instruction interface---this both improves the accuracy of the basic capabilities and advances the model from ``can do the task'' to ``can understand the task in various phrasings.'' The core of this stage is ``accurate'' and ``general'': raise accuracy on high-quality data while opening up the generalization of instruction following.

\subsubsection{Stage 4: Advanced-Capability Supervised Fine-Tuning}

\textbf{Objective.} After Stage 3 lays the instruction foundation and the accuracy of core capabilities, Stage 4 specifically strengthens real-world hard UI scenarios, advancing the model from ``can use simple interfaces'' to ``can read complex interfaces.'' This stage focuses on five categories of hard cases that general-purpose VLMs commonly fail: long interfaces, small elements, multi-step reasoning, cross-application generalization, and abnormal scenarios.

\textbf{Trainable parameters and hyperparameters.} Full-parameter end-to-end training, with a peak learning rate of 1e-5 (more conservative than Stage 3's 2e-5, to avoid overfitting on high-quality small-scale data), the ViT learning rate still at 1/10 of the LLM's, using a warmup + cosine-decay schedule~\cite{cosine_lr}. This stage accounts for about 5\% of the overall token budget, with a small sample size but a high density of learning signals.

\textbf{Data recipe.} We carefully select five categories of complex UI data:
\begin{itemize}[nosep]
    \item \textbf{Long-interface understanding ($\sim$30\%)}: scrolling screenshots spanning more than 3 screens, and deeply nested settings and feed pages; sources include in-house long screenshots, long documents from DocVQA~\cite{docvqa}, and synthetically stitched long screens, used to cultivate the model's grounding and summarization capabilities over very long visual sequences.
    \item \textbf{Small-element grounding ($\sim$20\%)}: extreme grounding tasks in which the target element occupies $<$1\% of the screen area, such as toolbar icons, toggle buttons, and status indicators, built by automatically mining the UI tree, to compensate for the model's perceptual blind spot for small-scale targets.
    \item \textbf{Multi-step reasoning VQA ($\sim$20\%)}: complex questions requiring 2--3 steps of visual reasoning to answer (e.g., ``What is the most expensive item currently in the shopping cart?''), with CoT synthesized by a stronger VLM and verified through rule-based consistency checks.
    \item \textbf{Cross-application generalization ($\sim$20\%)}: cross-application trajectory data sampled from GUIAct~\cite{guiact}, AndroidWorld~\cite{androidworld}, and AITW~\cite{aitw}, plus in-house screenshots covering mainstream domestic apps (WeChat, Alipay, Douyin, Meituan, etc.), to extend the model's robustness in the real-world ecosystem.
    \item \textbf{Abnormal scenarios ($\sim$10\%)}: boundary cases such as loading, error dialogs, permission requests, no network, and locked accounts, sampled from real-device logs and synthetic A/B interfaces, to train the model to recognize and gracefully handle non-standard states.
\end{itemize}

\textbf{Design and motivation.} Foundational SFT solves ``whether it can be done,'' while advanced SFT solves ``how well and how deeply it is done.'' We observe that general-purpose VLMs perform acceptably on simple UI tasks but degrade markedly once they encounter complex scenarios such as long interfaces, small elements, and multi-step reasoning, which is the main source of the gap between general-purpose VLMs and UI-specialized VLMs on benchmarks such as ScreenSpot~\cite{screenspot_v2} and complex-app grounding. By specifically introducing hard-case data, Stage 4 deepens these ``long-tail but high-frequency'' scenarios, enabling StepX-Edge to handle the UI vertical more finely within the same parameter tier.

\subsubsection{Stage 5: Reinforcement-Learning Alignment}

The RL stage advances the model from ``can do the task'' to ``reliable and easy to use'': on one hand, it continues to push up the accuracy ceiling of each task through verifiable rewards; on the other, it aligns output format and language style with real usage needs through preference-based rewards, while suppressing the two categories of hallucination that are prevalent in UI scenarios (misidentifying elements and fabricating undisplayed content). The overall design follows the \textbf{verifiable-reward + preference-reward + anti-hallucination} three-part framework of MiniCPM-V 4.5~\cite{minicpmv46}, with a series of engineering reinforcements for the dual constraints of a small model and the UI vertical.

\textbf{Optimization algorithm: GRPO + length-unbiased + EMA anchoring.} The main policy optimizer adopts GRPO (Group Relative Policy Optimization)~\cite{grpo}, sampling $K=8$ rollouts per prompt and estimating advantages via within-group relative rewards to dispense with a critic model, which is especially friendly to RL training under a 0.9B parameter count. We make corrections to the two major destabilization sources of the original GRPO: (1) \textbf{Token-level advantage normalization}---the log-prob difference of each rollout is first divided by the token count, then baseline-subtracted within the group, eliminating the implicit bias toward long answers (in UI mixed training, grounding is often only 3--5 tokens while screen VQA can reach dozens of tokens, and without normalization the imbalance is severe); (2) following MiniCPM-V 4.5, removing the KL and entropy penalties in favor of \textbf{periodic EMA reference anchoring}---every 200 steps the reference is updated from the policy with $\alpha=0.05$, balancing the efficiency of no-KL with long-horizon stability.

\textbf{RL data composition.} Four task categories are mixed for joint rollout: grounding ($\sim$40\%, SeeClick~\cite{seeclick} / AndroidWorld~\cite{androidworld}), OCR / text extraction ($\sim$25\%), screen VQA / understanding ($\sim$25\%, ScreenQA~\cite{screenqa} / Screen2Words~\cite{screen2words}), and multi-step UI interaction ($\sim$10\%, AITW~\cite{aitw} / AMEX~\cite{amex}).

\textbf{Reward quality control: from binary $\to$ dense.} We adopt differentiated dense rewards for different tasks:
\begin{itemize}[nosep]
    \item \textbf{Grounding: IoU + centerness dual-head reward}---$R_{\text{gnd}} = \mathrm{IoU} + \lambda \cdot \exp(-\|c_{\text{pred}} - c_{\text{gt}}\|^2/\sigma^2)$, where $c$ is the box center. The centerness term still provides usable gradients on extremely hard samples where IoU $\to$ 0, accelerating the convergence of small-element grounding.
    \item \textbf{OCR: field-level macro-F1}---computed after key-value matching on the JSON output, better fitting the UI-scenario reality of ``partially correct fields = partial reward'' than an overall Levenshtein distance.
    \item \textbf{VQA / screen understanding: RLPR probability reward}---following RLPR (Reference-based Log-Probability Reward)~\cite{rlpr}, using the token-generation probability of the reference answer under the policy model as a soft reward, circumventing rule brittleness.
    \item \textbf{Verifier ensemble (anti-Goodharting)}---all rewards requiring LLM scoring are independently scored by three teacher VLMs from different checkpoints and passed by majority vote, diluting the misjudgment caused by single-model hallucination.
\end{itemize}

\textbf{Reward shaping.} The final reward is a weighted combination of four components:
\begin{equation}
R = w_{\text{acc}} R_{\text{acc}} + w_{\text{fmt}} R_{\text{fmt}} + w_{\text{rep}} R_{\text{rep}} + w_{\text{rm}} R_{\text{rm}},
\end{equation}
corresponding respectively to task correctness, format compliance (coordinate format, JSON structural integrity, action-sequence validity), repetition penalty, and preference score, with typical weights $(0.5, 0.2, 0.1, 0.2)$.

\textbf{Hallucination suppression: an enhanced RLAIF-V.} The two prevalent categories of hallucination in UI scenarios are \textbf{misidentifying elements} (recognizing the ``share'' button as ``favorite'') and \textbf{fabricating undisplayed content} (describing fields that do not exist on the screen). On top of the atomic-proposition decomposition of RLAIF-V~\cite{rlaif_v}, we make two enhancements: (1) \textbf{Cross-VLM ensemble verification}---each proposition is independently scored by three heterogeneous teacher VLMs, and only $\geq$2 passes count as verified, reducing the Goodharting room of a single verifier; (2) \textbf{Positive-negative rewrite self-consistency}---each proposition is rewritten as ``$X$ exists'' $\leftrightarrow$ ``$X$ does not exist,'' and only when the positive and negative judgments are consistent is it included in the preference pair, removing samples on which the verifier itself is unstable. The filtered preference pairs are used to perform a second-order fine-tuning of the post-GRPO policy model via DPO (Direct Preference Optimization)~\cite{dpo}.

\textbf{Stability wrap-up.} Throughout RL, $\sim$5\% of Stage 3 SFT samples are mixed in for joint optimization to prevent catastrophic forgetting and the degradation of general instruction capability; the ViT continues the low learning-rate strategy from Stage 3 (1/10 of the LLM learning rate), participating in updates with extremely small steps to suppress the drift of visual representations during RL (a common latent side effect of RL-training VLMs).

\subsubsection{Layered Parameter-Unfreezing Strategy}

StepX-Curriculum adopts a layered parameter-unfreezing strategy: different modules enter training at different stages and use differentiated learning rates, which is a key design for the efficient transfer of small-parameter models:
\begin{itemize}[nosep]
    \item \textbf{ViT main layers}: frozen in Stages 1 and 2, unfrozen from Stage 3 at 1/10 of the LLM learning rate. Freezing the main layers protects the general visual knowledge from pretraining, while the low learning rate balances domain adaptation and knowledge retention.
    \item \textbf{Merger layer}: trainable throughout. As the high-level aggregator of visual features, its speed and quality of domain adaptation directly determine the ceiling of subsequent modality alignment; letting the Merger participate in training early enables rapid UI-domain adaptation of high-level visual features.
    \item \textbf{Connector}: trainable throughout. As the cross-modal bridge, it needs continuous adjustment throughout training to adapt to the ever-changing visual and language representations.
    \item \textbf{LLM}: trainable throughout. The language model adapts to multimodal inputs faster, and its early participation accelerates cross-modal semantic alignment.
\end{itemize}

\begin{takeawaybox}
\begin{itemize}[nosep]
    \item When the four UI tasks are jointly trained in a shared, constrained visual-representation space, they reinforce one another---grounding drives spatial encoding, OCR accumulates textual semantics, and they serve as mutual anchors, forming a positive feedback loop rather than interfering with one another.
    \item The key to the efficient transfer of small-parameter models is ``layered, staged'' unfreezing---ordering the training sequence and learning rates by each module's plasticity and knowledge value, letting the high-level aggregator adapt first and the low-level body fine-tune later.
\end{itemize}
\end{takeawaybox}

\subsection{On-Device Quantized Deployment Scheme}\label{sec:deploy}

The deployment pipeline is the key link taking an on-device VLM from ``benchmark-usable'' to ``real-device-usable,'' and it is also the link at which existing work most often remains at the simulation level. UI tasks are highly sensitive to precise text recognition and pixel-level grounding, yet the error introduced by low-bit quantization is amplified token by token during autoregressive decoding---relying solely on post-training quantization (PTQ) often incurs unacceptable accuracy loss. To this end, StepX-Edge follows the design philosophy of ``module-wise differentiated quantization + two-stage PTQ$\to$QAT'': targeting the respective computational characteristics and accuracy sensitivity of the vision encoder, modality adapter, and language decoder, we adopt differentiated quantization bit-widths and granularities; and on the Snapdragon 8 Gen5 platform, using the QAIRT inference stack, we complete the full real-device validation from model conversion, quantization, and on-device deployment to end-to-end inference.

\subsubsection{Module-Wise Differentiated Quantization Strategy}

The three modules differ significantly in their bottlenecks and accuracy sensitivity during on-device inference: the ViT performs only a single forward pass during inference and is sensitive to pixel-value perturbations; the main on-device bottleneck of the LLM is weight-access bandwidth; and the KV cache has a large impact on runtime memory footprint. Based on this observation, we assign differentiated quantization configurations to the different modules, as shown in Table~\ref{tab:quant}.

\begin{table}[!htbp]
\centering
\small
\caption{Module-wise quantization configuration of StepX-Edge.}\label{tab:quant}
\ra{1.25}
\begin{tabular}{@{}l c L{7.6cm}@{}}
\toprule
\textbf{Module}    & \textbf{Scheme}       & \textbf{Granularity and notes} \\
\midrule
ViT                & W8A16                 & Weights per-channel 8-bit / activations 16-bit \\
Connector          & W8A16                 & Weights per-channel 8-bit / activations 16-bit \\
LLM                & W4A16 + KV8           & Weights per-channel 4-bit / activations 16-bit / KV cache 8-bit (native) \\
Embedding          & 8-bit                 & 8-bit lookup table (LUT) \\
\bottomrule
\end{tabular}
\end{table}

\textbf{(1) ViT / Connector: W8A16.} The ViT and Connector use 8-bit per-channel weights, while activations are kept at 16-bit. This choice stems from a key engineering observation: in the residual stream of the last vision block (\texttt{blocks\_23}) of the vision encoder, there exist outlier activations orders of magnitude larger than ordinary signals (massive activation~\cite{massive_activations})---the magnitudes of a few channels can reach hundreds of times those of ordinary signals, and they are concentrated in a small number of fixed channels. Measurements show that this residual tensor spans roughly $[-244, 5987]$, whereas the magnitude of the vast majority of ordinary signals is only $O(10)$. If per-tensor 8-bit quantization~\cite{smoothquant} were applied to the activations of the vision pathway, a single outlier channel would saturate the quantization range of the entire tensor, compressing the vast majority of ordinary signals to near zero, and the visual features would be thoroughly distorted after the subsequent LayerNorm normalization. Therefore, the activations of the vision pathway uniformly use per-tensor 16-bit quantization.

\textbf{(2) LLM: W4A16 + KV8.} The main on-device cost of the LLM comes from the memory-access bandwidth of the weights, so we apply 4-bit per-channel quantization~\cite{gptq,awq} to the weights, shrinking the weight volume and access bandwidth to roughly one-quarter (a nearly 4$\times$ reduction relative to the bf16 baseline); among them, a very small number of layers highly sensitive to quantization error (\texttt{layer\_2.down\_proj}) retain 8-bit, safeguarding the accuracy of key layers at a negligible volume cost. Activations are kept at 16-bit to preserve the dynamic range of the intermediate representations in Attention and FFN; the KV cache uses 8-bit compression to further reduce memory footprint and bandwidth pressure in long-context scenarios. While significantly reducing on-device resource overhead, this combination keeps the impact of 4-bit quantization on autoregressive-generation quality within a range that QAT can compensate for.

\textbf{(3) Quantization granularity constrained by NPU operator support.} The choice of every quantization configuration is subject to the hard constraint of mobile-NPU operator support---standard full attention and mature operators are used throughout, avoiding schemes that require per-channel activation quantization or special operator support, ensuring that the quantized computation graph can be executed natively and correctly by the HTP backend. This follows the ``deployment-friendly first'' architectural philosophy of \S\ref{sec:arch} in this paper.

\subsubsection{Two-Stage PTQ$\to$QAT Quantization Pipeline}

We decompose quantization into a two-stage pipeline of ``one-shot PTQ freezing of the ViT, QAT compensation for the LLM,'' in which the two stages share the same frozen visual front-end, thereby decoupling training complexity while guaranteeing the consistency of visual representations.

\textbf{Stage one: ViT PTQ.} Post-training quantization is applied to the ViT and Connector. The calibration set is selected via a diversity-aware strategy---across dozens of heterogeneous data sources, k-means clustering is performed on low-level visual features and cross-source-balanced sampling is applied to construct 1024 calibration samples, so that the limited samples cover the visual distribution of UI scenarios as much as possible. The model performs a forward pass at a fixed input resolution of 768$\times$1344 (an on-device static shape, corresponding to a fixed number of visual tokens), gathers the activation ranges of each operator, and solves for the quantization parameters. After validating the visual-feature accuracy on the real device, the weights and quantization parameters of the ViT and Connector are frozen, serving as the fixed visual front-end for subsequent LLM training.

\textbf{Stage two: LLM QAT.} Freezing the above ViT + Connector (which only produces visual features via forward pass and does not participate in gradient updates), QAT is performed on the LLM alone. We first initialize the LLM's quantization parameters with a small calibration set of about 512 batches, then simultaneously update the language model's weights and quantization parameters via QAT---both are trained at the same learning rate ($2\times10^{-6}$), letting the model perceive and compensate for the quantization error of the 4-bit weights during training. Because the ViT and Connector parameters are fixed, the language pathway only needs to adapt to a stable visual-input distribution, making the training objective more focused and convergence more stable. Ultimately, the target of an accuracy loss within 1\% is achieved.

This two-stage decoupled design has two benefits: first, the ViT uses W8A16 quantization, for which PTQ alone can meet the accuracy requirement, obviating expensive vision-side QAT; second, the 4-bit quantization error of the LLM is compensated by QAT, and the same frozen visual front-end is reused throughout training, avoiding the mutual coupling and amplification of visual and language quantization errors.

\subsubsection{On-Device Deployment Pipeline and Runtime}

After quantization, the model is converted and compiled via the QAIRT~\cite{qairt} toolchain into a context binary executable on the Hexagon HTP~\cite{hexagon_htp}, and its execution is orchestrated by the Genie~\cite{genie} multimodal runtime.

\textbf{ViT subgraph.} The ViT and Connector are exported as a fixed-shape ONNX~\cite{onnx}, then compiled into an HTP context binary after QNN conversion and quantization. Thanks to the full-attention and standard-operator design, the vision subgraph executes natively on the HTP without special operator adaptation.

\textbf{LLM subgraph.} The LLM is exported as two static-shape computation graphs, for prefill and decode, which share the same 4-bit weights (weight sharing) and are compiled into an HTP context binary together with the native KV8 cache. The prefill graph is responsible for a one-shot parallel encoding of the prompt (including the visual embeddings), while the decode graph is responsible for token-by-token autoregressive generation; weight sharing avoids the memory waste of storing a separate copy of the weights for each graph.

\textbf{Runtime orchestration.} On the device, the Genie runtime executes a three-node pipeline: imageEncoder (the ViT subgraph, which encodes the image into visual embeddings) $\to$ lutEncoder (8-bit word-embedding lookup) $\to$ textGenerator (the LLM). Multimodal inputs are concatenated in the order [prefix text][image embeddings][suffix text], supporting the input form of ``UI screenshot + text instruction.'' The runtime library, HTP context binary, tokenizer, and word-embedding table are packaged uniformly as a device bundle, pushed to the real device once and kept resident in memory.

\begin{takeawaybox}
Module-wise differentiated quantization + the two-stage decoupling of ``freeze the ViT / compensate the LLM'' is the key to making ``accuracy loss $<$1\%'' and ``natively runnable on mobile NPUs'' hold simultaneously---the accuracy-sensitive vision side is frozen once at high bit-width, while the bandwidth-sensitive language side uses low bit-width + QAT compensation.
\end{takeawaybox}

\section{Experiments and Evaluation}


\subsection{Experimental Setup}

\textbf{Comparison baselines.} To ensure fairness of evaluation and reference value for on-device landing, we select six mainstream lightweight on-device multimodal models open-sourced within the past year as comparison baselines, with parameter counts ranging over 0.8B--2.3B: Qwen3.5-VL-0.8B~\cite{qwen35vl} (0.8B), InternVL3.5-1B~\cite{internvl35} (1B), MiniCPM-V 4.6~\cite{minicpmv46} (1.3B), LFM2.5-VL-1.6B~\cite{lfm2vl} (1.6B), Qwen3-VL-2B-Instruct~\cite{qwen3vl} (2B), and Gemma4-E2B-it~\cite{gemma4} (2.3B).

\textbf{Evaluation benchmarks.} We measure the model's UI-understanding capability along three dimensions:
\begin{itemize}[nosep]
    \item \textbf{OCR text recognition}: OCRBench v1~\cite{ocrbench} and OCRBench v2~\cite{ocrbenchv2} (reported separately for Chinese and English), evaluating general text recognition and document parsing;
    \item \textbf{Referring grounding}: RefCOCO~\cite{refcoco}, measuring pixel-level element referring and grounding accuracy in general scenarios;
    \item \textbf{GUI understanding}: ScreenSpot v2~\cite{screenspot_v2} (Mobile subset) and ScreenQA~\cite{screenqa}. The former is a mobile GUI-specific benchmark covering element referring and interaction grounding on real app interfaces; the latter evaluates the understanding and question answering of screen content.
\end{itemize}

\textbf{Implementation details.} All benchmark scores are obtained on a single A100 GPU; each model uses its official open-source weights and default inference configuration, with input resolution and prompt templates fully aligned with StepX-Edge. The on-device deployment experiments target only StepX-Edge and are validated on real hardware on the Qualcomm Snapdragon 8 Gen5 (SM8850) platform, with a quantization configuration of ViT+Connector W8A16 and LLM W4A16 + KV Cache 8-bit.

\textbf{Evaluation protocol.} To ensure cross-model comparability, each benchmark adopts the following unified conventions:
\begin{itemize}[nosep]
    \item \textbf{OCRBench v1} reports the official total score (out of 1000); \textbf{OCRBench v2} is reported separately on the English (en) and Chinese (cn) subsets, taking the average of the official metrics for each subset.
    \item \textbf{RefCOCO} uses the val split, counting a hit when the IoU between the predicted box and the ground-truth box is $\geq$ 0.5, and reports accuracy (Acc@0.5).
    \item \textbf{ScreenSpot v2 (Mobile)} reports the click-grounding accuracy on the mobile subset (a hit occurs when the predicted point falls within the target element's bounding box).
    \item \textbf{ScreenQA} uses a generative QA convention, taking the token-level F1 between the model output and the reference answer as the metric.
\end{itemize}

All baseline scores are obtained by us using the official open-source weights, under input resolutions and prompt templates fully consistent with StepX-Edge, to eliminate incomparability arising from differences in evaluation configuration.

\subsection{Main Results}

\begin{figure}[!t]
    \centering
    \includegraphics[width=\linewidth]{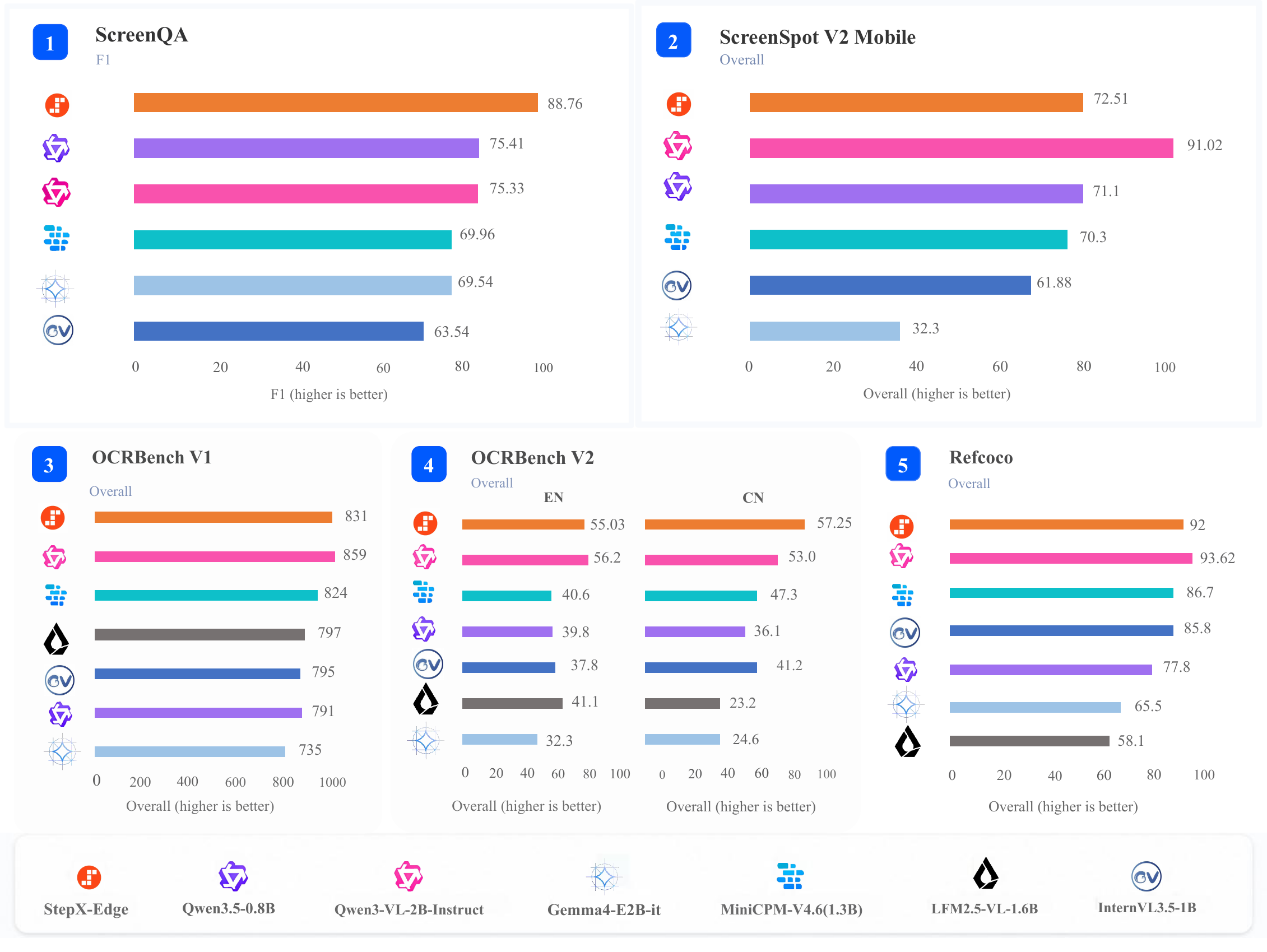}
    \caption{Comparison of StepX-Edge with mainstream on-device VLMs across UI-understanding benchmarks. With 0.9B parameters, StepX-Edge achieves the highest scores overall on ScreenQA and Chinese OCRBench v2, and matches 1.3B--2.3B general-purpose VLMs on benchmarks such as RefCOCO and OCRBench v1 with a smaller parameter count.}
    \label{fig:main}
\end{figure}

The comprehensive experimental results are shown in Figure~\ref{fig:main}. With only 0.9B parameters, StepX-Edge achieves the strongest overall UI-understanding performance in the $\leq$1B tier; it surpasses all baselines (including 2B--2.3B ones) on screen question answering and Chinese OCR, and matches 1.3B--2.3B general-purpose VLMs on general grounding and OCR with a smaller parameter count.

\textbf{Screen question answering (ScreenQA).} StepX-Edge attains an F1 of 88.76, the highest among all baselines, leading the runner-up Qwen3.5-VL-0.8B (75.41) by more than 13 points, and also markedly exceeding the 2B--2.3B Qwen3-VL-2B (75.33) and Gemma4-E2B-it (69.54). The model can fully understand the hierarchical structure and functional logic of interfaces, and its robustness in boundary scenarios---multi-component coordination, permission dialogs, abnormal interfaces---is clearly superior to that of general-purpose small models.

\textbf{Chinese OCR (OCRBench v2 cn).} StepX-Edge ranks first among all baselines with 57.25, surpassing the runner-up Qwen3-VL-2B (53.0) and MiniCPM-V 4.6 (47.3). On the English subset (en) it attains 55.03, leading the same tier ($\leq$1B) and essentially on par with 2B-class models (Qwen3-VL-2B and LFM2.5-VL at 56.2). For highly structured images such as invoices and tables, the model demonstrates strong fine-grained field-extraction capability, able to output format-compliant JSON structures.

\textbf{General OCR (OCRBench v1).} StepX-Edge attains 831, outperforming the same-tier Qwen3.5-VL-0.8B (791) and the 1.3B MiniCPM-V 4.6 (824), and approaching the 2B Qwen3-VL-2B (859)---reaching the level of a 2B-class model with about 0.9B parameters.

\textbf{Referring grounding (RefCOCO).} StepX-Edge attains a grounding accuracy of 92.0\%, surpassing same-tier models and the 1.3B MiniCPM-V 4.6 (86.7) and InternVL3.5-1B (85.8), and approaching the 2B Qwen3-VL-2B (93.62). This confirms the improvement in spatial-encoding capability brought by multi-task mutually-reinforcing training---attaining pixel-level grounding accuracy close to the strongest baseline with far fewer parameters than 2B.

\textbf{Mobile GUI grounding (ScreenSpot v2 Mobile).} StepX-Edge attains 72.51, leading among same-tier ($\leq$1B) models (vs. InternVL3.5-1B's 61.88) and comparable to 1.3B/1.6B models (MiniCPM-V 4.6 at 70.3). On this benchmark, the 2B Qwen3-VL-2B leads with 91.02, reflecting that interaction grounding on complex mobile app interfaces still has room to improve with scale, and is a key direction for subsequent optimization.

\textbf{Overall.} General-purpose VLMs pursue broad ``jack-of-all-trades'' coverage, whereas StepX-Edge, through specialized optimization for the vertical scenario, surpasses all baselines (including 2B+) on screen understanding and Chinese OCR with 0.9B parameters, and matches larger models on grounding and general OCR with fewer parameters---achieving ``smaller and more precise'' in the UI niche.

\subsubsection{Discussion of Design Effectiveness}

The above results reveal a clear pattern: \textbf{StepX-Edge's advantages concentrate on UI scenarios that most rely on ``fine-grained perception + spatial encoding + cross-task synergy,'' rather than on sheer scale-stacking.} We map this pattern to the design choices of Chapter~2 as follows:

\textbf{The overall-first results on screen question answering and Chinese OCR are consistent with multi-task mutually-reinforcing training.} ScreenQA (88.76) leads the runner-up by more than 13 points, and Chinese OCRBench v2 (57.25) surpasses all larger models. Both task categories simultaneously rely on ``reading text precisely'' and ``understanding the position and semantics of text within the interface''---precisely the scenarios that directly benefit from the OCR--grounding mutual reinforcement described in \S\ref{sec:training}: grounding-driven spatial encoding helps the model correctly segment text lines and locate fields, while the textual semantics accumulated by OCR feed back into the understanding of interface content. That a 0.9B model can surpass 2B-class general-purpose VLMs here shows that the gains from synergy outweigh the gap in raw parameter scale.

\textbf{Matching larger models on grounding and general OCR is consistent with ULVE's spatial-perception optimization.} RefCOCO (92.0) approaches the strongest 2B baseline, and OCRBench v1 (831) approaches the 2B level, reflecting that the adaptive tiling and positional-encoding recalibration for extreme aspect ratios in \S\ref{sec:arch} effectively narrows the gap to larger models on pixel-level grounding and dense-text recognition.

\textbf{No overall drop on fine-grained tasks, consistent with PDP's information preservation.} The OCR and grounding tasks that rely on fine-grained visual information show no significant degradation common to small models, consistent with the design goal of PDP's ``refine before reduction, adapt after reduction'' to mitigate the information bottleneck in \S\ref{sec:arch}.

\textbf{The weaknesses also point to clear optimization directions.} ScreenSpot v2 Mobile (72.51) still has a gap to the 2B model (91.02), concentrated mainly on interaction grounding for complex real app interfaces; such complex scenarios are precisely the data type emphasized in the later-curriculum Stage 4 (advanced-capability SFT), and are the main focus for subsequent improvement.

\subsection{GPU Server-Side Runtime Comparison}

\begin{figure}[!t]
    \centering
    \includegraphics[width=1\linewidth]{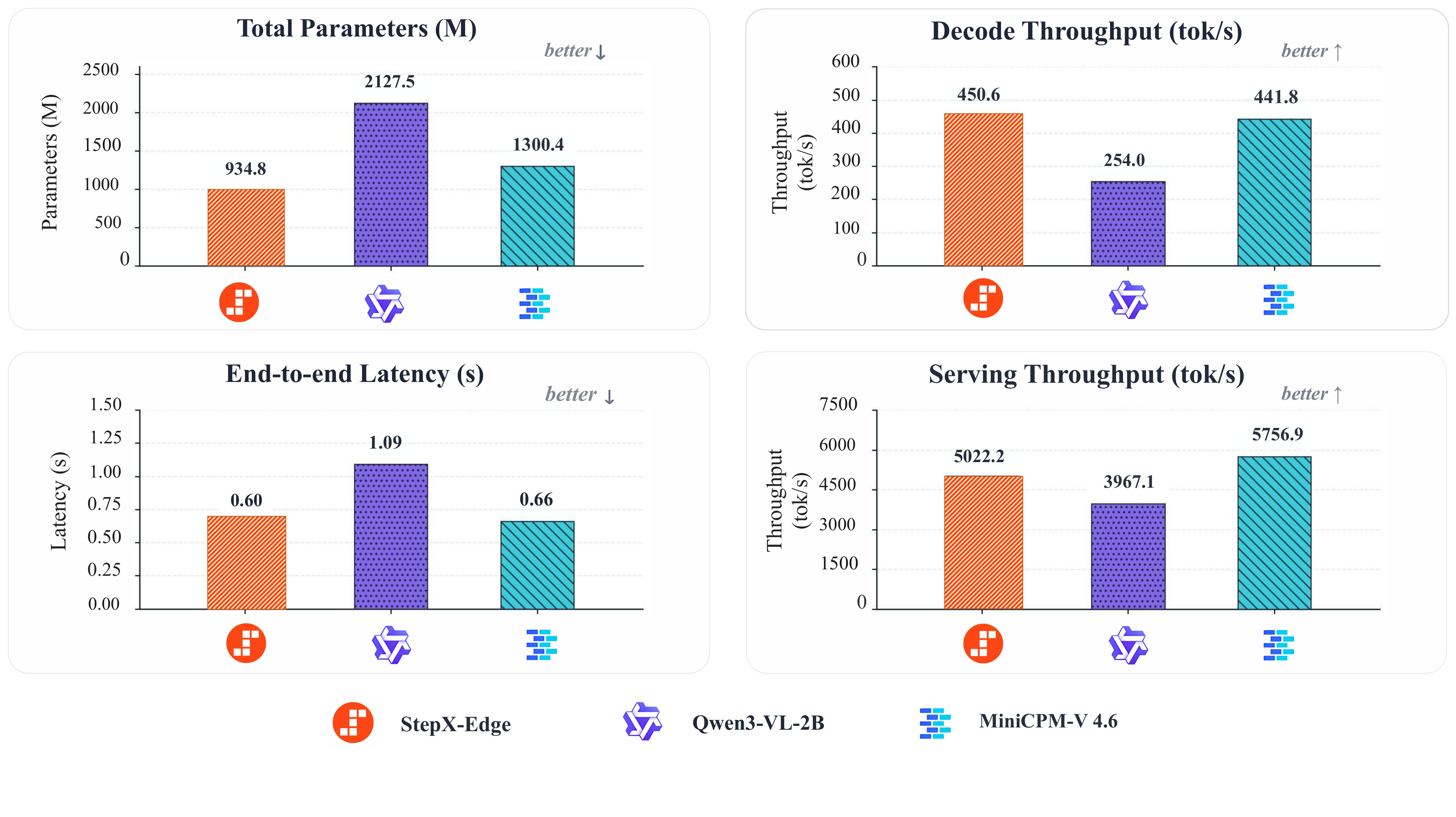}
    \caption{GPU server-side runtime comparison of StepX-Edge with mainstream on-device VLMs (NVIDIA A800, vLLM 0.22.1, bf16, CUDA Graph enabled): the four metrics of parameter count, decode throughput, single-image end-to-end latency, and batched-serving throughput ($B=32$).}
    \label{fig:runtime}
\end{figure}

We measure the runtime efficiency of StepX-Edge relative to comparable models in the server-side setting. To ensure comparability, on the same NVIDIA A800 GPU and using the \textbf{same version of vLLM (0.22.1)~\cite{vllm}}, we conduct a horizontal comparison of StepX-Edge, Qwen3-VL-2B-Instruct, and MiniCPM-V 4.6: all three are loaded via their respective \textbf{native vLLM model classes} (rather than a generic inference backend, to align operator fusion and scheduling paths), with CUDA Graph uniformly enabled, bf16 precision, the same 768$\times$1344 UI screenshot and instruction input, greedy decoding, and max\_new\_tokens=272. We report three server-side metrics: single-request decode throughput (linearly fitted via a two-point method to isolate prefill and visual-encoding overhead), single-image end-to-end latency, and batched-serving throughput under 32-way concurrency ($B=32$). The results are shown in Table~\ref{tab:runtime} and Figure~\ref{fig:runtime}.

\begin{table}[!htbp]
\centering
\small
\caption{GPU server-side runtime comparison of StepX-Edge with mainstream on-device VLMs. Setup: NVIDIA A800, vLLM 0.22.1, bf16, CUDA Graph enabled; single-image 768$\times$1344 UI input, greedy decoding, \texttt{max\_new\_tokens=272}; each model is loaded via its native vLLM model class.}\label{tab:runtime}
\ra{1.3}
\setlength{\tabcolsep}{6pt}
\begin{tabular}{@{}l c C{2.6cm} C{2.6cm} C{2.8cm}@{}}
\toprule
\textbf{Model} & \textbf{Params} & \textbf{Decode tput.\ (tok/s)} & \textbf{Single-image E2E (s)} & \textbf{Serving tput.\ ($B{=}32$, tok/s)} \\
\midrule
Qwen3-VL-2B-Instruct        & 2B    & 254.0          & 1.09          & 3967 \\
MiniCPM-V 4.6               & 1.3B  & 441.8          & 0.66          & \textbf{5757} \\
\midrule
\textbf{StepX-Edge (ours)}  & 0.9B  & \textbf{450.6} & \textbf{0.60} & 5022 \\
\bottomrule
\end{tabular}
\end{table}

\textbf{Decode throughput and single-image latency.} With the smallest parameter count (0.9B), StepX-Edge attains the highest decode throughput (450.6 tok/s) and the lowest single-image end-to-end latency (0.60 s): its decode speed is about 1.8$\times$ that of the 2B Qwen3-VL-2B (254.0 tok/s), and its end-to-end latency is about 55\% of the latter's. This is consistent with the difference in parameter scale---a smaller language decoder means lower per-token memory access and computation; it also confirms that the ``full attention + standard operators'' design of \S\ref{sec:arch} can be fully optimized by a mature inference stack, so the model pays no speed penalty for its deployment-friendly architectural trade-offs.

\textbf{Batched-serving throughput.} Under $B{=}32$ concurrency, StepX-Edge reaches 5022 tok/s, markedly higher than Qwen3-VL-2B (3967 tok/s); MiniCPM-V 4.6, at 5757 tok/s, is slightly higher than StepX-Edge. This difference stems mainly from the number of visual tokens: MiniCPM's slice-based visual encoding produces only about 444 visual tokens on this input, whereas both StepX-Edge and Qwen3-VL produce 1008---fewer visual tokens reduce prefill computation and KV-cache pressure, yielding higher throughput under high concurrency. It should be noted that the visual-token resolution directly relates to fine-grained UI-perception capability (see \S\ref{sec:arch}), and StepX-Edge's retention of a higher visual-token count is one prerequisite for its lead on OCR and grounding; there is a clear ``resolution--throughput'' trade-off between the two. Even so, StepX-Edge's batched throughput remains clearly superior to that of Qwen3-VL-2B, which has the same visual-token count and twice the parameters.

\textbf{Summary.} Under equal server-side conditions, StepX-Edge achieves the best decode throughput and single-image latency, together with competitive batched throughput, with less than half the parameters of its competitors.

\subsection{On-Device Inference Efficiency}

We complete an end-to-end validation on a Snapdragon 8 Gen5 device: for real UI/document screenshots, the model can output coherent and accurate OCR and understanding results, validating the functional completeness of the ``ViT $\to$ W4A16/KV8 LLM'' pipeline on real hardware. In terms of system performance (single image, mobile-resolution input, about 272 output tokens), the measured on-device latencies and resource footprints are shown in Table~\ref{tab:perf}.

\begin{table}[!htbp]
\centering
\small
\caption{On-device inference performance of StepX-Edge (measured on a Snapdragon 8 Gen5 device).}
\label{tab:perf}
\ra{1.25}
\begin{tabular}{@{}l l L{6.8cm}@{}}
\toprule
\textbf{Metric}            & \textbf{Measured value} & \textbf{Convention / notes} \\
\midrule
ViT (HTP execute)          & $\sim$0.59 s     & 1008 visual tokens, multiple HVX threads \\
Prefill                    & $\sim$0.25 s     & $\sim$1048 prompt tokens, $\sim$4200 tok/s \\
TTFT                       & $\sim$0.84 s     & Visual encoding + prefill \\
Decode                     & $\sim$98 tok/s   & Native KV8, token-by-token autoregression \\
Single-image E2E           & $\sim$3.6 s      & Visual encoding + prefill + 272-token decode \\
Peak runtime memory        & $\sim$1.4 GB     & Weight mmap + activation/KV cache + runtime \\
\bottomrule
\end{tabular}
\end{table}

The combination of module-wise differentiated quantization, 4-bit weight sharing, and the KV8 cache keeps on-device inference latency and memory footprint within the acceptable range of mainstream mobile devices while preserving UI-understanding accuracy: a time-to-first-token of about 0.84 s, decoding at about 98 tok/s, and a peak memory of about 1.4 GB. With about 0.9B parameters (only about 39\% of Gemma4-E2B-it and about 69\% of MiniCPM-V 4.6), StepX-Edge closes the loop from benchmark accuracy to real-device usability.

\section{Conclusion}

This paper presents StepX-Edge, a 0.9B-parameter on-device vision--language model designed specifically for mobile UI understanding. Targeting the resource constraints of on-device deployment and the specialized needs of UI scenarios, we systematically optimize along four dimensions---architecture design, data engineering, training strategy, and quantized deployment: we propose the UI-Aware Layered Visual Encoding strategy (ULVE) and the Progressive Dimensionality Projection (PDP) modality adapter; we build, mainly from public datasets, a large-scale training corpus of about 10.5M samples covering the four core capabilities together with a full-pipeline data quality-control process; we design the five-stage progressive StepX-Curriculum training framework based on the discovery of the mutual-reinforcement effect among tasks; and we complete the real-device landing validation of two-stage PTQ$\to$QAT quantization on the Snapdragon 8 Gen5 platform.

Experimental results show that, with only 0.9B parameters, StepX-Edge achieves the strongest overall UI-understanding performance in the $\leq$1B tier: it attains the highest scores among all baselines (including larger 2B--2.3B models) on ScreenQA and Chinese OCRBench v2, and matches 1.3B--2.3B general-purpose VLMs on benchmarks such as RefCOCO and OCRBench v1 with a smaller parameter count. At the same time, on a Snapdragon 8 Gen5 device, it achieves a decode speed of about 98 tok/s, a peak runtime memory of about 1.4 GB, and a time-to-first-token of about 0.84 s, with smooth and usable end-to-end inference. Across the dual dimensions of accuracy and efficiency, StepX-Edge validates the advantage of ``specialized optimization for vertical scenarios + a systematic training methodology'' over the coarse pretraining of general-purpose small models.

Future work will proceed in three directions: first, further optimizing the quantization strategy and on-device inference framework to compress model size and latency and to adapt to mobile devices with even lower compute; second, refining the later-curriculum advanced-capability SFT and reinforcement-learning alignment stages, and extending UI-Agent capabilities to support more complex interaction scenarios such as multi-step operations and cross-application task execution; and third, open-sourcing the training data, the complete training pipeline, and the quantized deployment scheme, offering the community a reproducible recipe for building on-device VLMs.

\clearpage
\bibliographystyle{unsrtnat}
\bibliography{references}

\clearpage
\appendix

\section{Qualitative Case Studies}\label{sec:cases}

To intuitively present the actual performance of StepX-Edge in UI scenarios, this appendix selects one Chinese and one English example from each of the three task categories---OCR, UI screen understanding, and VQA (six real screenshots in total)---to show the model's actual inputs and outputs. All outputs are generated on the Snapdragon 8 Gen5 device with the W4A16+KV8 quantized model.

\subsection{OCR Text Recognition}\label{sec:case:ocr}

\noindent
\begin{caseframe}
\centering
\includegraphics[width=0.42\linewidth,height=10cm,keepaspectratio]{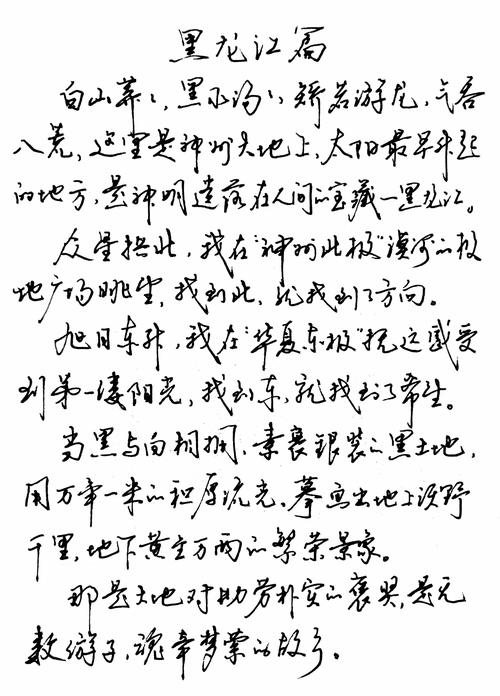}
\par\vspace{0.8em}
\begin{minipage}{\linewidth}
    \begin{promptbox}
    \textbf{Prompt.} Please recognize all the text content on this page of handwritten notes, preserving the original paragraph breaks and punctuation.
    \end{promptbox}
    \vspace{0.5em}
    \noindent\textbf{\small StepX-Edge:}
    \begin{responsebox}
    \textbf{黑龙江篇}\\[0.2em]
    白山莽，黑水汤，矫若游龙，气吞八荒，这里是神州大地上，太阳最早升起的地方，是神明遗落在人间的宝藏 —— 黑龙江。\\[0.2em]
    众星拱北，我在``神州北极''漠河的极地广场眺望，找到北，就找到了方向。\\[0.2em]
    旭日东升，我在``华夏东极''抚远感受到第一缕阳光，我向东，就找到了希望。\\[0.2em]
    当黑与白相拥，素裹银装的黑土地，用万年一米的积厚流光，摹画出地上沃野千里，地下黄金万两的繁荣景象。\\[0.2em]
    那是大地对勤劳朴实的褒奖，是无数游子，魂牵梦萦的故乡。
    \end{responsebox}
\end{minipage}
\end{caseframe}
\captionof{figure}{OCR case (Chinese handwriting scenario): a handwritten manuscript of the essay \emph{Heilongjiang}. The model correctly recognizes the title, paragraph breaks, and all text, including proper nouns such as ``神州北极'' (Northern Pole of China) and ``华夏东极'' (Eastern Pole of China).}\label{fig:case_ocr_cn}

\clearpage

\noindent
\begin{caseframe}
\centering
\includegraphics[width=0.42\linewidth,height=10cm,keepaspectratio]{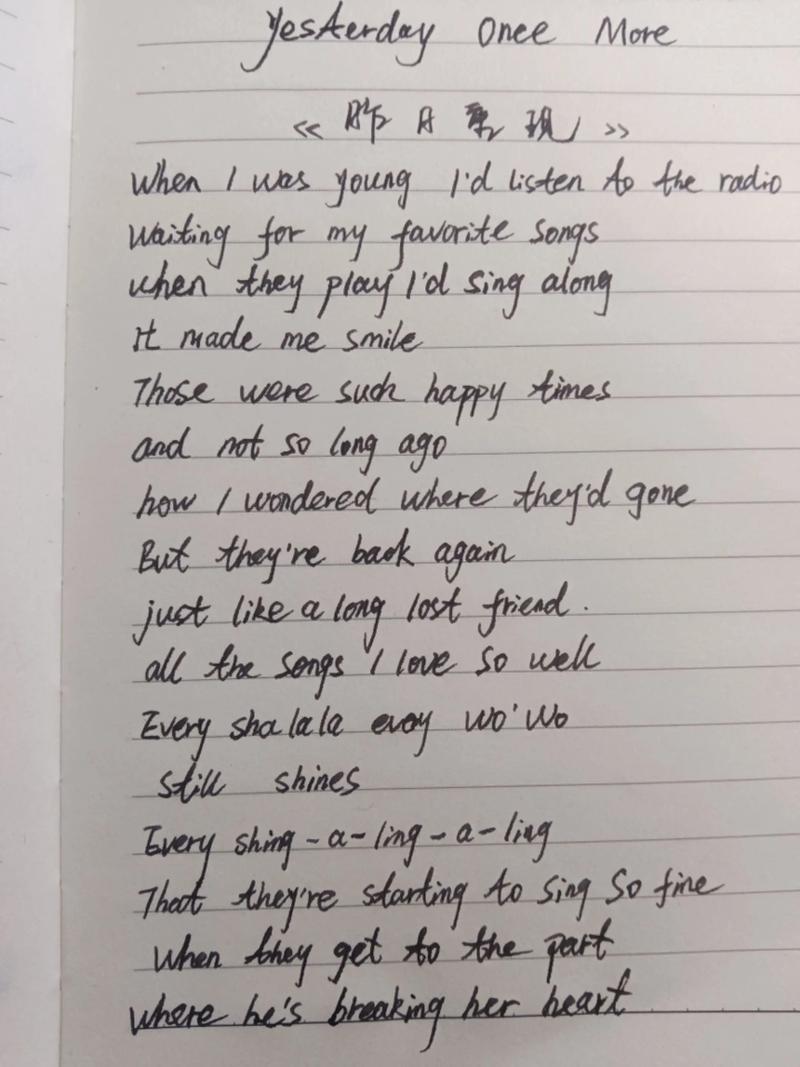}
\par\vspace{0.8em}
\begin{minipage}{\linewidth}
    \begin{promptbox}
    \textbf{Prompt.} Transcribe all handwritten text on this page, preserving the title, subtitle and line breaks.
    \end{promptbox}
    \vspace{0.5em}
    \noindent\textbf{\small StepX-Edge:}
    \begin{responsebox}
    \textbf{Yesterday Once More}\\
    《昨日重现》\\[0.2em]
    when I was young I'd listen to the radio\\
    waiting for my favorite songs\\
    when they play I'd sing along\\
    It made me smile\\
    Those were such happy times\\
    and not so long ago\\
    how I wondered where they'd gone\\
    But they're back again\\
    just like a long lost friend.\\
    all the songs I love so well\\
    Every sha la la every wo'wo\\
    still shines\\
    Every shing-a-ling-a-ling\\
    That they're starting to sing so fine\\
    When they get to the part\\
    where he's breaking her heart
    \end{responsebox}
\end{minipage}
\end{caseframe}
\captionof{figure}{OCR case (English handwriting scenario): a handwritten manuscript of the lyrics of \emph{Yesterday Once More}. The model transcribes the title, the Chinese translated name, and all lyrics line by line, and, drawing on semantic context, restores the handwritten strokes into a well-formed song title and lyrics.}\label{fig:case_ocr_en}

\clearpage

\subsection{UI Screen Understanding}\label{sec:case:ui}

\noindent
\begin{caseframe}
\centering
\includegraphics[width=0.42\linewidth,height=10cm,keepaspectratio]{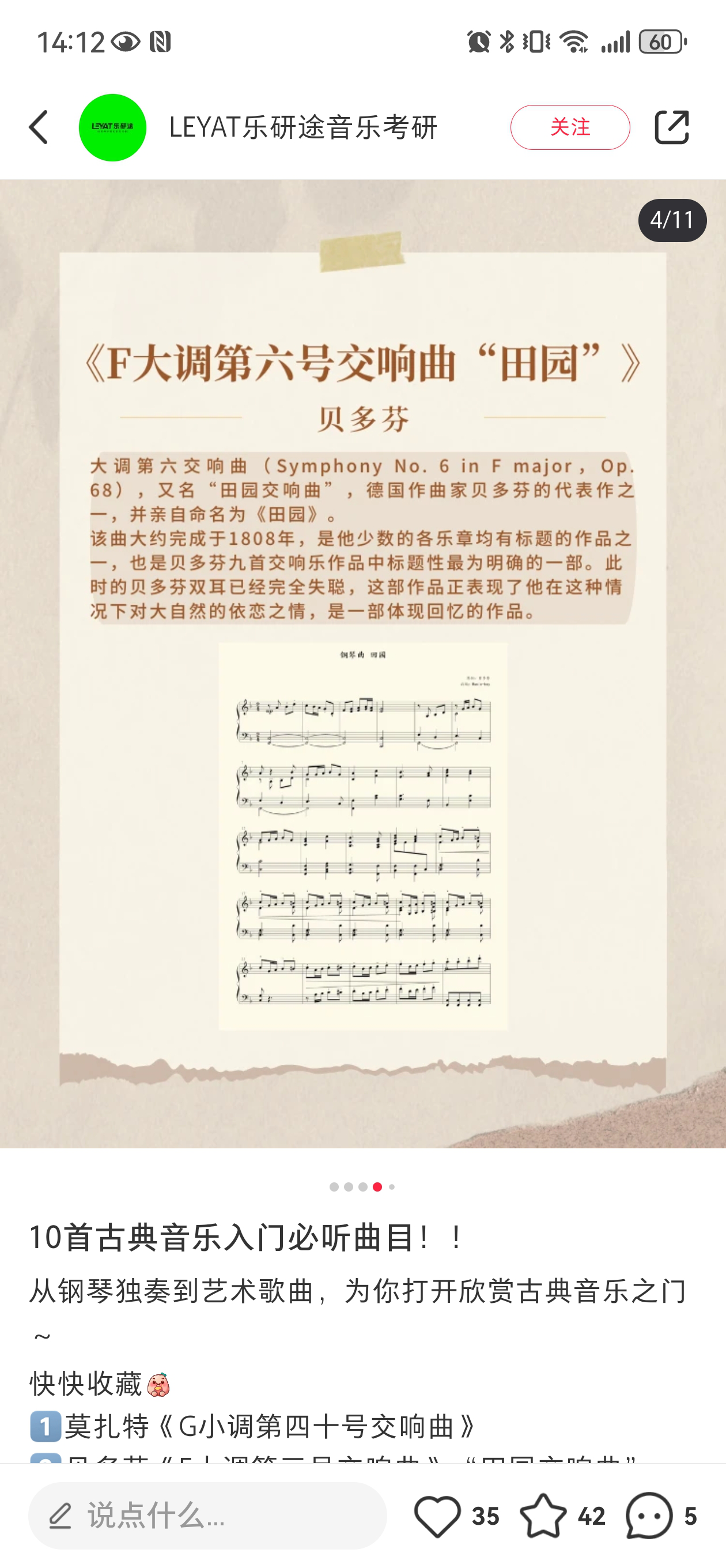}
\par\vspace{0.8em}
\begin{minipage}{\linewidth}
    \begin{promptbox}
    \textbf{Prompt.} 请按功能分组，描述这个内容分享 App 页面上展示了哪些信息。
    \end{promptbox}
    \vspace{0.5em}
    \noindent\textbf{\small StepX-Edge:}
    \begin{responsebox}
    这是一个内容分享（笔记类）App 的帖子详情页，主要由三个语义区块组成：\\[0.2em]
    \textbf{（1）顶部导航栏}：左侧为返回按钮，中间是带头像的作者``LEYAT 乐研途音乐考研''，右侧有``关注''按钮和分享按钮。\\
    \textbf{（2）图片卡片}：为 11 张图中的第 4 张（右上角标注``4/11''），介绍贝多芬的《F 大调第六号交响曲``田园''（Op.~68）》，配有一段文字说明和一页钢琴谱。\\
    \textbf{（3）文案与互动}：标题为``10 首古典音乐入门必听曲目！！''，下方是曲目推荐列表；底部栏包含评论输入框，以及点赞 35、收藏 42、评论 5 的计数。\\[0.2em]
    卡片下方的圆点指示器表示当前图片在轮播中的位置。
    \end{responsebox}
\end{minipage}
\end{caseframe}
\captionof{figure}{UI screen-understanding case (Chinese scenario): a post-detail page of a note-sharing app (a classical-music recommendation post). The model correctly outputs the semantic sections and the interaction counts.}\label{fig:case_ui_cn}

\clearpage

\noindent
\begin{caseframe}
\centering
\includegraphics[width=0.42\linewidth,height=10cm,keepaspectratio]{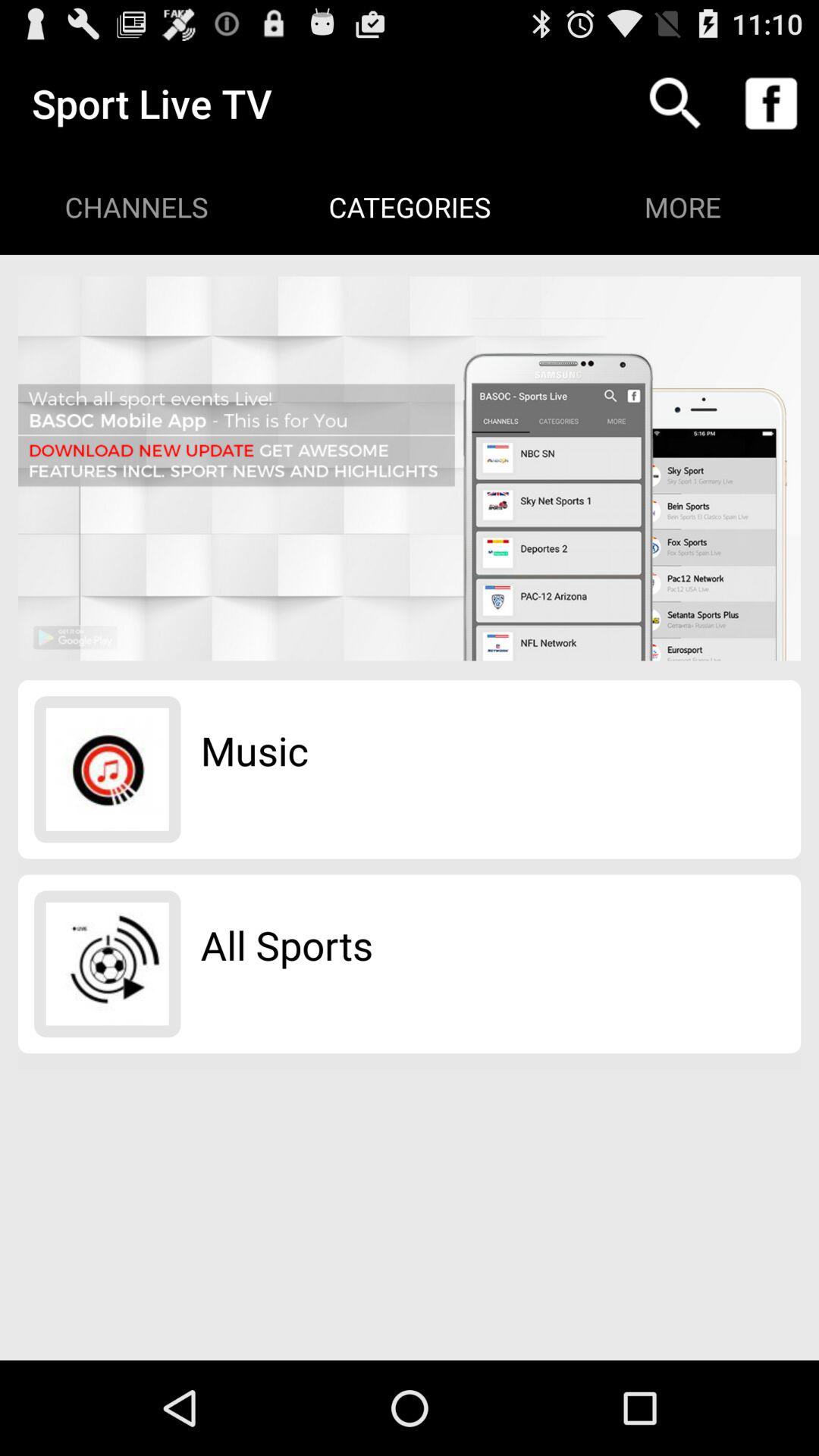}
\par\vspace{0.8em}
\begin{minipage}{\linewidth}
    \begin{promptbox}
    \textbf{Prompt.} Describe the layout and content of this app screen.
    \end{promptbox}
    \vspace{0.5em}
    \noindent\textbf{\small StepX-Edge:}
    \begin{responsebox}
    This is the "Sport Live TV" app, currently on the \textbf{Categories} tab (three top tabs: Channels / Categories / More).\\[0.2em]
    \textbf{Top banner}: promotion for the BASOC mobile app, asking users to download a new update for sports news and highlights.\\[0.2em]
    \textbf{Category list}:
    \begin{itemize}[nosep,leftmargin=1.2em]
        \item Music --- audio-note icon
        \item All Sports --- live-signal icon
    \end{itemize}
    Top-right corner has a search button and a Facebook shortcut icon.
    \end{responsebox}
\end{minipage}
\end{caseframe}
\captionof{figure}{UI screen-understanding case (English scenario): the Categories page of the Sport Live TV app. The model recognizes the tab bar, the banner, and the category entries.}\label{fig:case_ui_en}

\clearpage

\subsection{Visual Question Answering (VQA)}\label{sec:case:vqa}

\noindent
\begin{caseframe}
\centering
\includegraphics[width=0.42\linewidth,height=10cm,keepaspectratio]{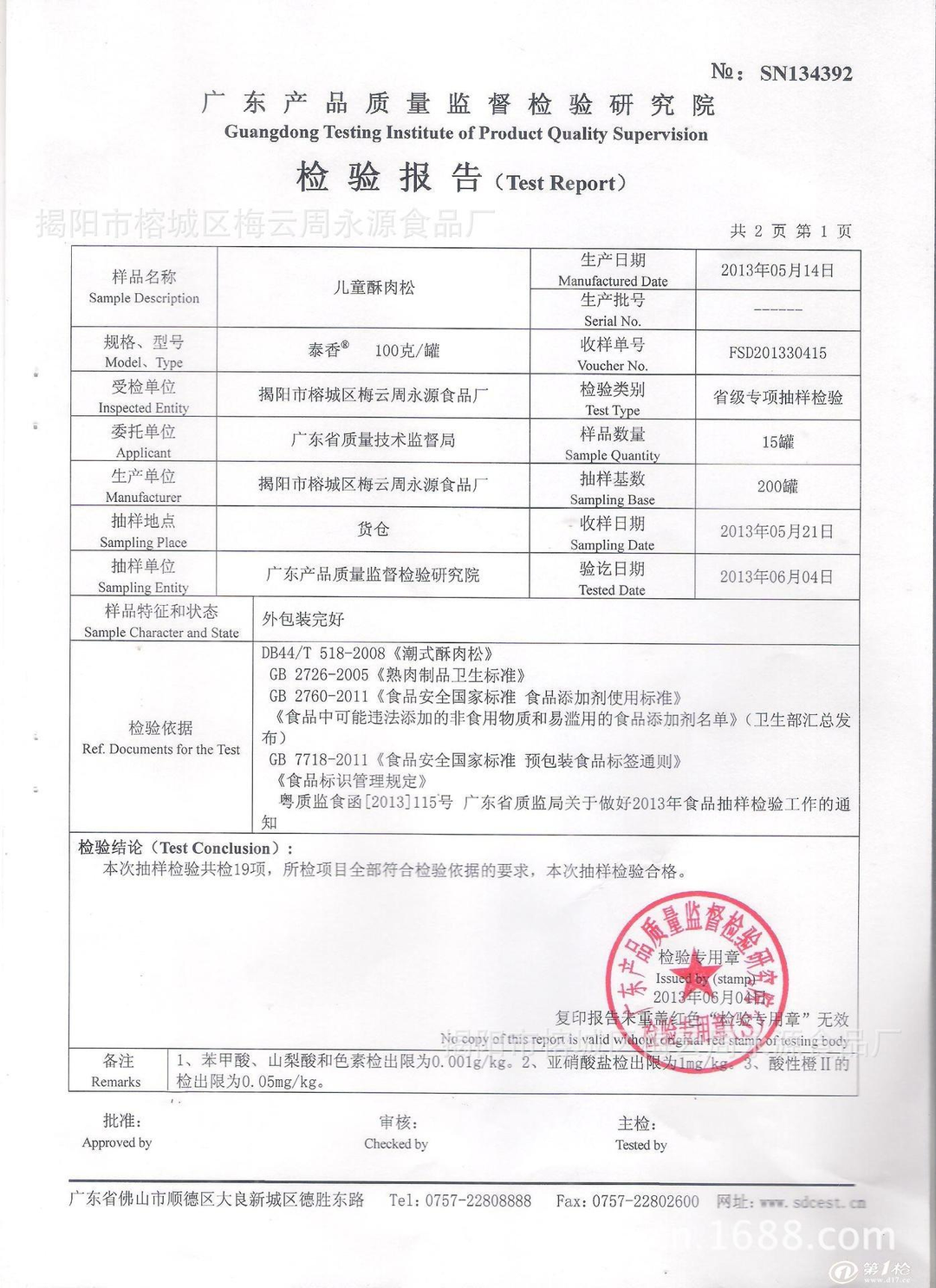}
\par\vspace{0.8em}
\begin{minipage}{\linewidth}
    \begin{promptbox}
    \textbf{Prompt.} 这份检验报告从收样日期到验讫日期一共经过了多少天？本次检验依据了几项标准或文件？该样品是否合格？
    \end{promptbox}
    \vspace{0.5em}
    \noindent\textbf{\small StepX-Edge:}
    \begin{responsebox}
    根据这份广东产品质量监督检验研究院出具的《检验报告》（编号 SN134392），回答需要综合报告中的多个字段：\\[0.2em]
    \textbf{经过天数}：收样日期为 2013 年 05 月 21 日，验讫日期为 2013 年 06 月 04 日，两者相差 \textbf{14 天}。\\
    \textbf{检验依据数量}：``检验依据''一栏共列出 \textbf{7} 项——DB44/T 518-2008、GB 2726-2005、GB 2760-2011、《食品中可能违法添加的非食用物质和易滥用的食品添加剂名单》、GB 7718-2011、《食品标识管理规定》，以及粤质监食函[2013]115 号通知。\\
    \textbf{是否合格}：合格。检验结论显示本次共检 19 项，所检项目全部符合检验依据的要求，故本次抽样检验\textbf{合格}。
    \end{responsebox}
\end{minipage}
\end{caseframe}
\captionof{figure}{VQA case (Chinese scenario): a product quality test report. The question requires multi-step reasoning---computing the day gap between two dates, counting the reference-document entries, and reading the pass/fail conclusion---which the model answers correctly.}\label{fig:case_vqa_cn}

\clearpage

\noindent
\begin{caseframe}
\centering
\includegraphics[width=0.42\linewidth,height=10cm,keepaspectratio]{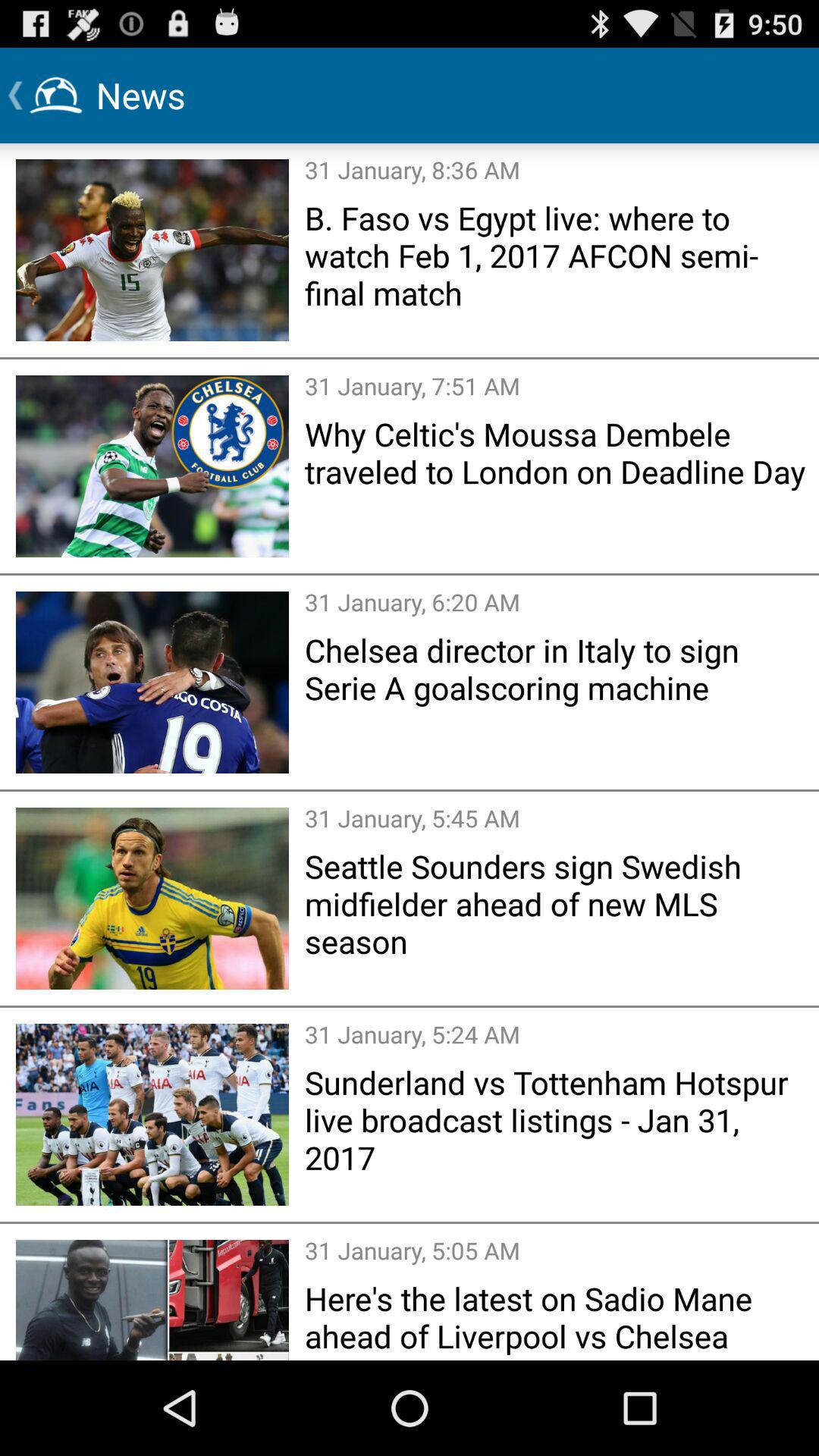}
\par\vspace{0.8em}
\begin{minipage}{\linewidth}
    \begin{promptbox}
    \textbf{Prompt.} How long does this recipe take, how many servings, and what are the ingredients?
    \end{promptbox}
    \vspace{0.5em}
    \noindent\textbf{\small StepX-Edge:}
    \begin{responsebox}
    Recipe: \textbf{Protein Vanilla Mango Smoothie}\\[0.2em]
    \textbf{Preparation time}: 5 min\\
    \textbf{Servings}: 4\\
    \textbf{Ingredients (3 items)}:
    \begin{itemize}[nosep,leftmargin=1.2em]
        \item 2 (400\,g) tins of mango pulp
        \item 100\,g yoghurt
        \item 60\,ml milk
    \end{itemize}
    \end{responsebox}
\end{minipage}
\end{caseframe}
\captionof{figure}{VQA case (English scenario): a recipe detail page. The model aggregates three categories of structured information---duration, servings, and ingredients.}\label{fig:case_vqa_en}

\end{document}